# Pattern Recognition of Ozone-Depleting Substance Exports in Global Trade Data

MUHAMMAD SUKRI BIN RAMLI
Asia School of Business
Kuala Lumpur, Malaysia
Email: m.binramli@sloan.mit.edu

**Abstract**

New methods are needed to monitor environmental treaties, like the Montreal Protocol, by reviewing large, complex customs datasets. This paper introduces a framework using unsupervised machine learning to systematically detect suspicious trade patterns and highlight activities for review. Our methodology, applied to 100,000 trade records, combines several ML techniques. Unsupervised Clustering (K-Means) discovers natural trade archetypes based on shipment value and weight. Anomaly Detection (Isolation Forest and IQR) identifies rare "mega-trades" and shipments with commercially unusual price-per-kilogram values. This is supplemented by Heuristic Flagging to find tactics like vague shipment descriptions. These layers are combined into a priority score, which successfully identified 1,351 price outliers and 1,288 high-priority shipments for customs review. A key finding is that high-priority commodities show a different and more valuable value-to-weight ratio than general goods. This was validated using Explainable AI (SHAP), which confirmed vague descriptions and high value as the most significant risk predictors. The model's sensitivity was validated by its detection of a massive spike in "mega-trades" in early 2021, correlating directly with the real-world regulatory impact of the US AIM Act. This work presents a repeatable unsupervised learning pipeline to turn raw trade data into prioritized, usable intelligence for regulatory groups.

## 1. Introduction

The Montreal Protocol stands as a landmark environmental agreement, yet its enforcement is profoundly challenged by the complexities of modern global trade, where illicit activities are concealed within millions of legitimate transactions. This issue is a component of a much broader environmental crime crisis that poses a direct threat to sustainable development and undermines international treaties (Nellemann et al., 2012). Illicit trade in Ozone-Depleting Substances (ODS) is a documented and ongoing problem, with investigative reports frequently exposing sophisticated smuggling networks that exploit regulatory loopholes and deliberately misclassify goods to evade detection (Environmental Investigation Agency, 2021). The core problem addressed by this research is the difficulty of identifying these illicit ODS shipments, which are hidden within large, poorly formatted, and heterogeneous customs datasets aggregated by the UN Comtrade system. Traditional methods, such as simple rule-based checks, are insufficient for discovering novel or concealed smuggling tactics. This deficit necessitates an unsupervised learning approach capable of performing advanced pattern recognition without relying on prior labelling of suspicious activities, which is critical in a domain where labelled training data is often obsolete or entirely unavailable (Bolton & Hand, 2002). While the application of data mining to detect customs fraud has shown promise (Camacho & Disch, 2014), the inherent noise and heterogeneity of real-world trade data demand a more robust, multi-modal framework.

**Figure 1. System Dynamics of Trade Misclassification and Regulatory Enforcement.**

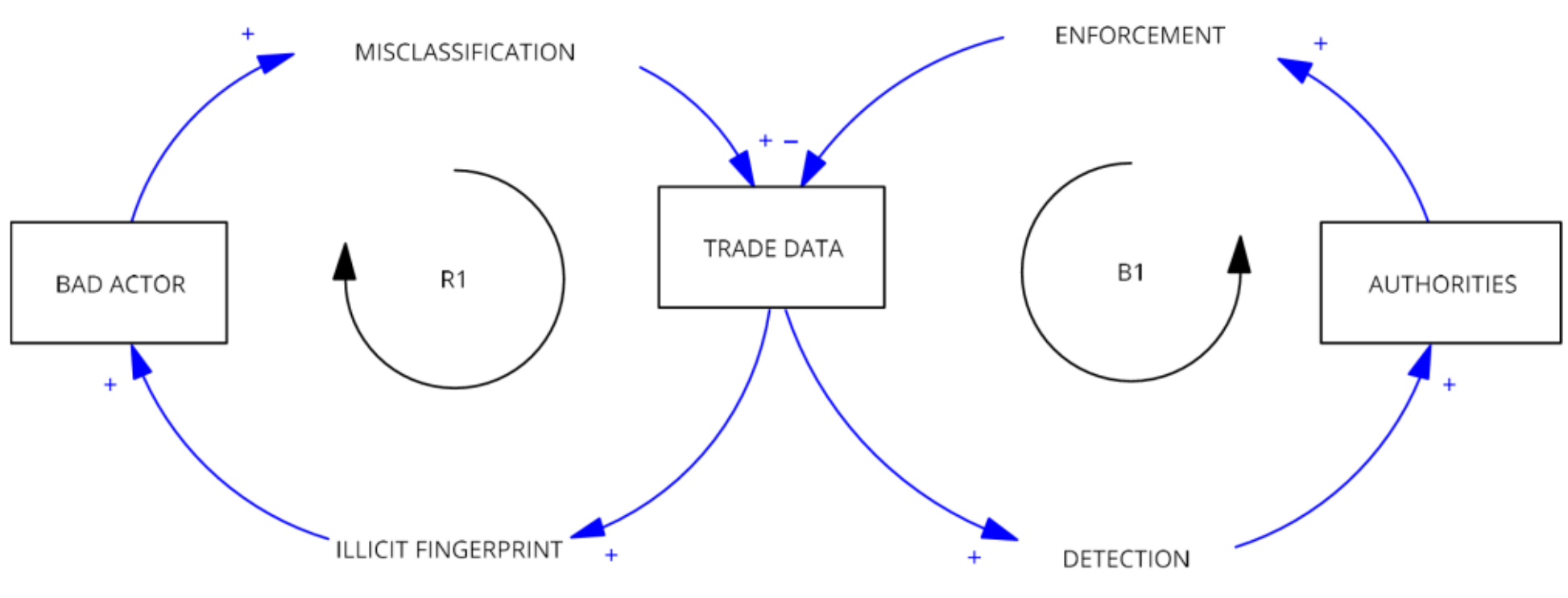

Source: Processed by Author (2025)

This research conceptualizes the relationship between illicit traders and enforcement through two competing feedback loops (Figure 1): a reinforcing loop (R1) of illicit activity and a balancing loop (B1) of detection. The R1 loop is driven by bad actors who engage in MISCLASSIFICATION of ODS shipments, using tactics like deliberately vague descriptions (e.g.,

'mixtures', 'halogenated derivatives'). Paradoxically, this activity creates an ILLICIT FINGERPRINT-a set of recognizable patterns within the trade data, such as commercially illogical price-per-kilogram values and anomalous value-to-weight ratios. Our multi-modal unsupervised learning framework is engineered to activate the B1 loop by precisely targeting these patterns. For instance, the Heuristic Flagging module (Layer 3) directly counters MISCLASSIFICATION by algorithmically flagging these vague descriptions. Simultaneously, the Unsupervised Clustering (Layer 1) and Commercial Anomaly Detection (Layer 2) modules detect the resulting ILLICIT FINGERPRINT by isolating the anomalous value-to-weight archetypes and commercially illogical price outliers. The framework then translates these combined findings into a single composite risk score. This final output is the DETECTION: a package of actionable intelligence (such as hotspot reports and prioritized case files) that enables AUTHORITIES to apply targeted ENFORCEMENT and disrupt the illicit activity. The principal contribution of this paper is, therefore, a comprehensive framework that combines unsupervised clustering with multiple anomaly detection techniques to create a prioritized, actionable intelligence package from raw, noisy trade data. This work demonstrates a methodological shift from manual inspection to automated forensic pattern analysis, transforming raw data into actionable intelligence, a core tenet of the forensic intelligence paradigm (Leloup, 2017). We aim to design, implement, and validate a multi-modal unsupervised pattern recognition pipeline that can automatically discover underlying trade structures, detect statistically and commercially anomalous transactions, and profile and prioritize high-risk trade networks for targeted investigation.

## 2. Methodology

The core of this research is a multi-layered pattern recognition engine designed to analyse trade data from different perspectives, recognizing distinct types of anomalous patterns. This structure follows an ensemble methodology; the combination of diverse unsupervised models has been shown to improve the stability and reliability of anomaly detection (Zimek et al., 2014). This unsupervised approach is fundamentally designed to first discover the baseline structures and "normal" archetypes within the data before identifying significant deviations from those established norms. This multi-modal strategy is rooted in the understanding that different classes of anomalies require distinct detection techniques, a concept well-established in the literature (Chandola et al., 2009). The full forensic pipeline consists of data preparation, a three-layered detection engine, risk prioritization, and advanced forensic overlays.

### 2.1 Robust Data Ingestion and Preparation

The foundational challenge in analyzing the dataset of 100,000 trade entries, sourced from the United Nations Comtrade database, was the integrity of the data itself. Initial attempts to load the data using standard, high-level parsing functions failed due to severe formatting inconsistencies, a common issue that renders many real-world forensic datasets unusable for automated analysis. To overcome this, we first engineered a resilient data ingestion framework. This critical data parsing solution was implemented using a robust, line-by-line ingestion method with Python's native csv module, which successfully handled formatting errors and transformed the raw, unusable data into a structured format suitable for analysis. This process, which aligns with the principles of tidy data (Wickham, 2014), positions the custom parsing script not as a simple preliminary step, but as a foundational and replicable component of the forensic pipeline, designed to ensure analytical integrity from the outset.

### 2.2 Multi-Layered Unsupervised Pattern Recognition Engine

The detection engine integrates three distinct forensic layers. The first layer focuses on Trade Archetype Discovery using unsupervised clustering. We applied the K-Means algorithm (with n=4 clusters) to the log-transformed primaryValue (the shipment's monetary value) and netwgt (its net weight) features. This unsupervised technique automatically partitions the data to establish a baseline of "normal" trade patterns, a foundational step in unsupervised learning for discovering inherent groupings in a feature space (Hastie et al., 2009). These partitions were interpreted as distinct trade archetypes, such as "Low Weight / High Value (Specialty)" and "High Weight / Low Value (Bulk)". This segmentation is visualized in the quadrant analysis presented in Figure 2, which plots all trade records clustered into four patterns based on the log of net weight and primary value. The second layer performs Commercial Anomaly Detection through a price-per-unit analysis. To flag commercially illogical patterns, a price_per_kg metric was engineered for each transaction. The Interquartile Range (IQR) method, a robust statistical form of unsupervised outlier detection originating from exploratory data analysis (Tukey, 1977), was then applied on a per-HS-code basis. This non-parametric technique was deliberately chosen for its robustness to extreme outliers, allowing the engine to effectively identify price anomalies relative to their specific commodity peer group rather than the entire dataset. Figure 3, "Price-per-kg Distribution for ODS HS Codes," illustrates this method's efficacy, displaying the price distributions on a log scale and clearly marking the individual transactions flagged as outliers by the IQR method. The third layer introduces Heuristic Flagging to capture domain-specific knowledge. This layer was designed to flag entries with vague commodity descriptions using keywords known to be associated with intentional obfuscation, such as 'mixtures', 'halogenated', and 'n.e.c.' (not elsewhere specified or included). This vague_desc flag serves as a critical method of feature engineering (Zheng & Casari, 2018), transforming qualitative, domain-specific knowledge about smuggling tactics into a machine-readable feature that a purely statistical model might otherwise miss.

**Figure 2: Anomaly Detection Quadrant Analysis of related Ozone Depleting Substance.**

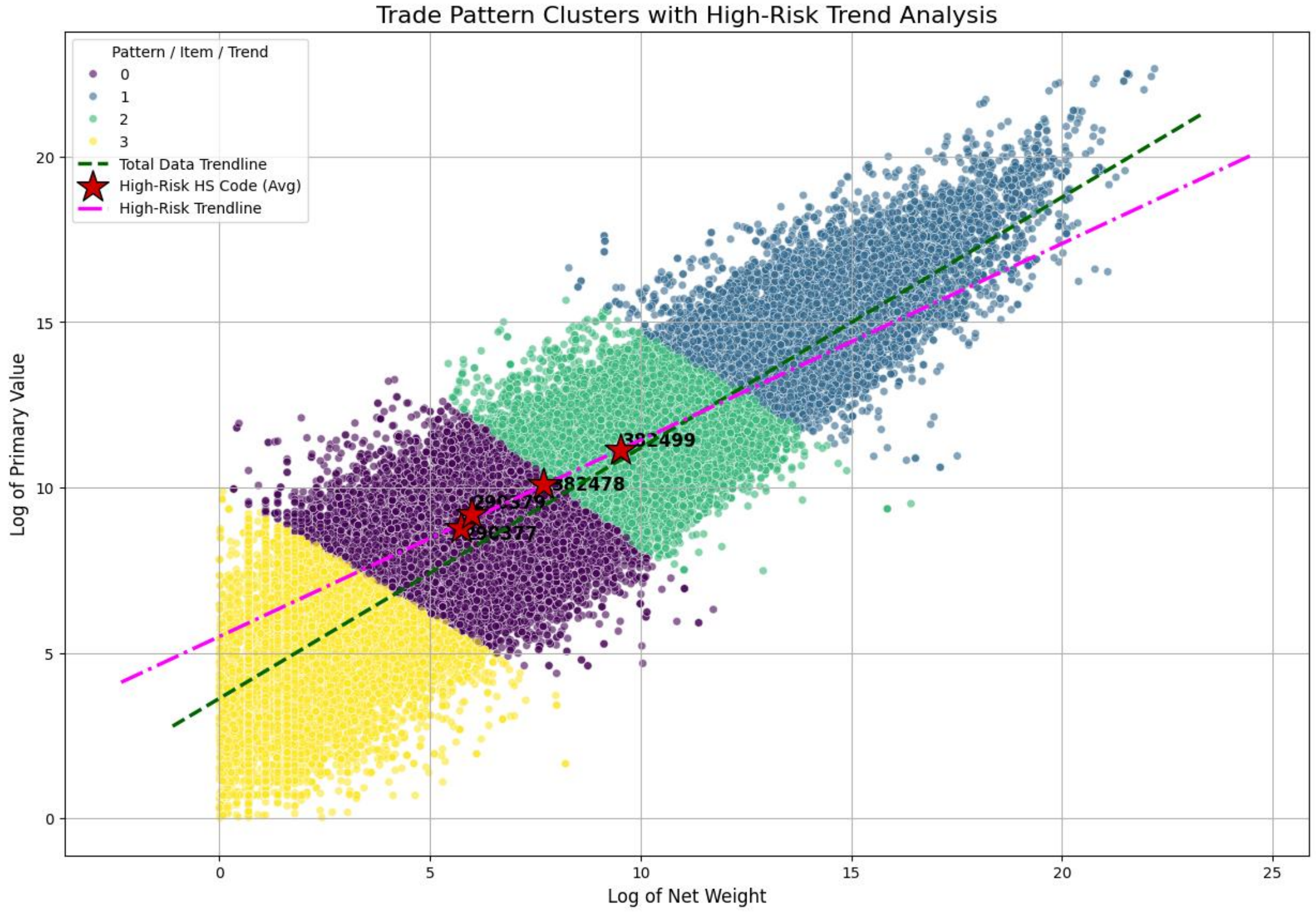

Source: Processed by Author (2025) using UN Comtrade Data (2020–2022)

The scatter plot on Figure 2 visualizes trade data to identify patterns and high-risk items by plotting the "Log of Primary Value" (Y-axis) against the "Log of Net Weight" (X-axis) for various trade shipments. This analysis helps identify the value-to-weight ratio of goods. The data points are grouped into four distinct clusters, each represented by a different color: Cluster 0 (purple), Cluster 1 (blue-grey), Cluster 2 (green), and Cluster 3 (yellow). Two trendlines are overlaid on the plot: a dashed green line representing the "Total Data Trendline" (the average for all data) and a pink dash-dot line representing the "High-Risk Trendline," which shows a different value-to-weight relationship. The key indicators on the chart are three red stars, which mark the average plot position for specific "High-Risk HS Codes." The codes 290379 and 382478 are shown falling squarely within the purple Cluster 0, while code 382499 is located within the green Cluster 2. This demonstrates that the clustering algorithm successfully segments the trade data, and it suggests that these specific clusters (0 and 2) are strongly associated with high-risk or anomalous trade activity.

**Figure 3. Price-per-kg Distribution for ODS HS Codes.**

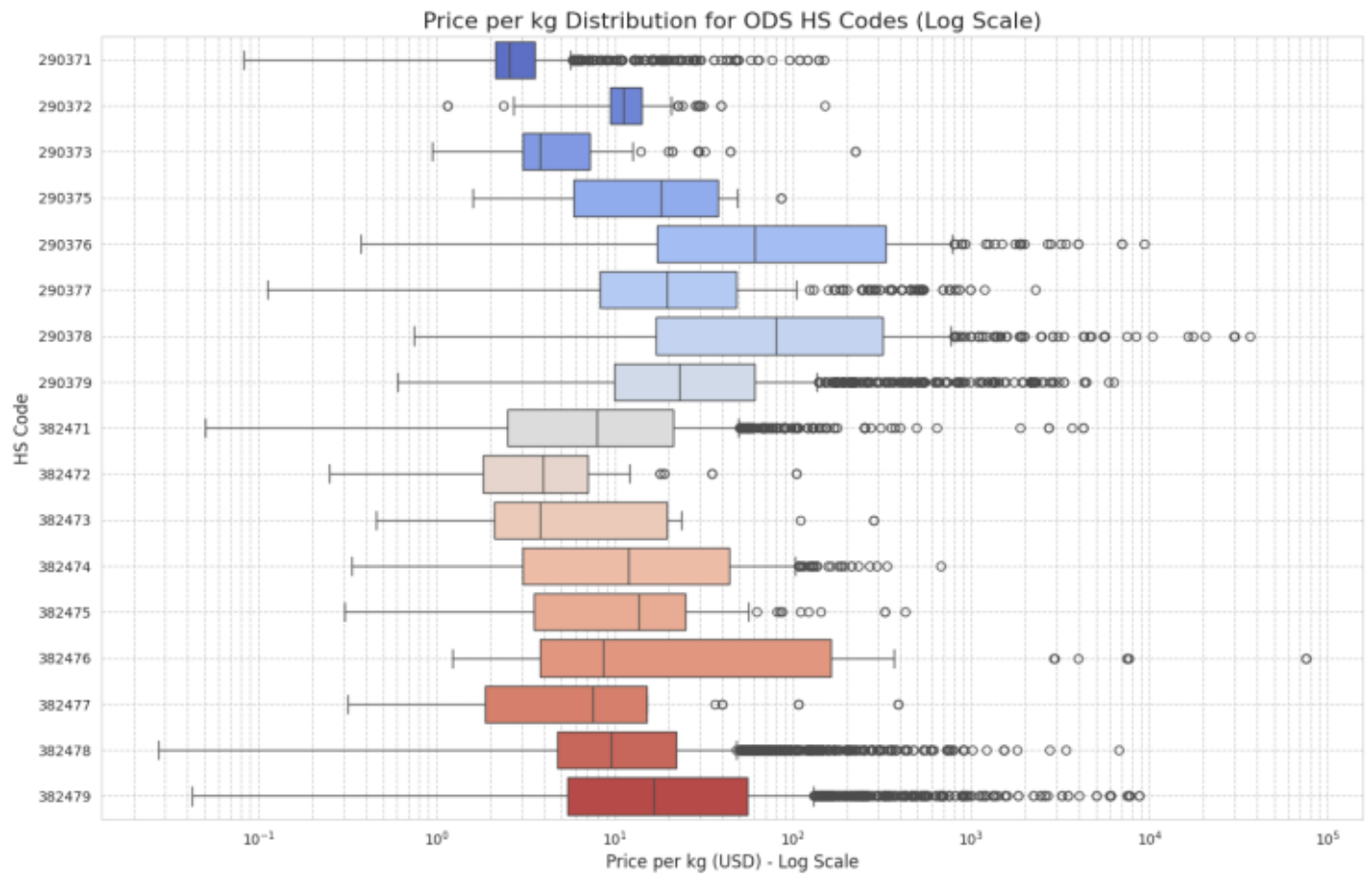

Source: Processed by Author (2025) using UN Comtrade Data (2020–2022)

### 2.3. Anomaly Triage and Risk Prioritization

Once patterns were detected by the various layers, the framework moved to anomaly triage and risk prioritization. The price anomalies recognized in Layer 2 were first triaged into two distinct queues: anomalies with a netwgt below 1.0 kg were routed to a "Data Quality Review" queue, while those with significant weight were prioritized for a "Customs Review" queue. To synthesize all findings from the different layers into a single, prioritized ranking, a final, composite risk score was developed. This score was formulated as a weighted scorecard, a standard methodology for risk stratification and prioritization (Siddiqi, 2017), which combined multiple factors such as the price anomaly and value score (e.g., price_score * 0.7 + value_score * 0.3). This composite score allows for a more nuanced ranking of suspicion than any single flag could provide alone. The box plot in Figure 3 visualizes the statistical distribution of prices for various Harmonized System (HS) codes, which are classifications for traded goods. The vertical Y-axis lists the specific HS codes (from 290371 down to 382479), while the horizontal X-axis represents the "Price per kg (USD)" on a logarithmic scale. Each horizontal entry on the chart is a box-and-whisker plot, summarizing the range of prices recorded for that single HS code. The colored box represents the interquartile range (IQR), which contains the middle 50% of all transactions. The vertical line inside each box indicates the median price. The "whiskers" (the horizontal lines extending from the box) show the typical spread of the data, while the individual circles represent outlier transactions with prices that fall significantly outside this main range. The use of a log scale on the X-axis highlights the immense variation in pricing, both between different codes and within the same code. Some codes, such as 290371, show a very tight and consistent price range. In contrast, other codes, particularly 382478 and 382479 at the bottom, display a much wider price distribution (a larger box) and a vast number of high-price outliers. This suggests that the goods classified under these latter codes are subject to extreme price volatility or contain many anomalous transactions. The chart also applies a blue-to-red color gradient to visually group the different HS codes.

### 2.4. Advanced Forensic Overlays for Deeper Insight

Finally, to validate these findings and extract deeper structural insights, the framework incorporates advanced forensic overlays. First, to discover structural patterns of collaboration, the trade network was modeled using NetworkX. The Louvain method, a highly efficient algorithm for fast community detection in large networks (Blondel et al., 2008), was applied to identify distinct "trade blocs". The application of network analysis to identify such cohesive subgroups is a cornerstone of modern criminology for understanding the structure of illicit enterprises (Malm & Bichler, 2011). Second, an Explainable AI (XAI) analysis using SHAP (SHapley Additive exPlanations) was conducted to interpret the patterns learned by a Random Forest model (Breiman, 2001) trained to predict the composite risk scores. This method provides a unified approach to interpreting model predictions by assigning each feature an importance value (Lundberg & Lee, 2017). Third, a geospatial risk mapping using a choropleth map (seen in Figure 13) was generated to visualize the geographic patterns of risk, providing a global overview of high-risk hotspots, a standard practice in criminal justice for providing a strategic overview of risk concentration (Chainey & Ratcliffe, 2013).

## 3. Results and Findings

The application of the multi-modal framework yielded several significant findings, revealing distinct, recognizable patterns that differentiate high-risk trades from general commerce. These results not only quantify the scale of suspicious activity but also validate the core hypotheses of the detection methodology.

### 3.1. Discovery of Illicit Trade Archetypes and Patterns

A key discovery emerged from the Trendline Divergence Analysis, which compared the value-to-weight relationship of different trade types. As visualized in Figure 2, the K-Means clustering successfully partitioned the dataset into four primary trade archetypes (e.g., "Bulk," "Specialty") based on their log-transformed value and weight. Building on this, a regression analysis revealed two distinct trendlines, as illustrated in Figure 4, " High-Risk vs. Total Data Trendline Comparison of Specific HS Code of ODS". The trendline for known high-risk Harmonized System (HS) codes (shown in magenta) exhibited a significantly steeper slope than the trendline for the general dataset (shown in dark green). This recognized pattern of divergence is a novel finding, indicating that high-risk goods possess a fundamentally different, and more valuable, value-to-weight ratio than general commerce. This provides a robust, quantifiable indicator for suspicion, moving beyond simple outlier detection to identifying a contextual anomaly in the economic behavior of specific commodities.

This scatter plot in figure 4 plots the "Log of Primary Value" (Y-axis) against the "Log of Net Weight" (X-axis). The chart visualizes two sets of data: a large, dense cluster of light blue data points that represents the main body of trade, and a much smaller set of yellow data points. Two different trendlines are overlaid: a green dashed line showing the average trend for the main (light blue) data, and a pink dash-dot line representing the specific "High-Risk Trendline." The central focus of the chart is a cluster of red stars, which are labeled as the "High-Risk HS Code (Avg)." These markers represent the average value-to-weight position for specific high-risk codes, including 290377, 290379, 382478, and 382499. These high-risk codes are grouped tightly together, closely following both the total data trendline and the high-risk trendline. The plot is divided into four quadrants by dashed red lines, which are centered on this high-risk cluster. These quadrants are labelled to categorize the data based on weight (W) and value (V): "Low W / Low V" (bottom left), "Low W / High V" (top left), "High W / Low V" (bottom right), and "High W / High V" (top right). This division highlights that the yellow data points are concentrated in the bottom-left and top-right corners, indicating they are outliers with either very low value and weight or very high value and weight, falling far from the primary data cloud.

The framework's commercial anomaly detection layer, which applied the IQR method to price_per_kg values within each HS code (visualized in Figure 3), successfully profiled and triaged these anomalous patterns. This layer identified a total of 1,351 price anomalies representing commercially illogical transactions. After a triage process that filtered out trades with insignificant weight (e.g., < 1.0 kg), 1,288 of these anomalies were deemed significant and forwarded to a "Customs Review" queue. A deep dive into these flagged transactions identified a specific high-volume trade corridor as a recurring hotspot. Figure 5 provides a detailed histogram of the anomalous prices on this specific route, clearly showing a high frequency of transactions with commercially illogical price-per-kg values that warrant further investigation.

**Figure 4: High-Risk vs. Total Data Trendline Comparison of Specific HS Code of ODS.**

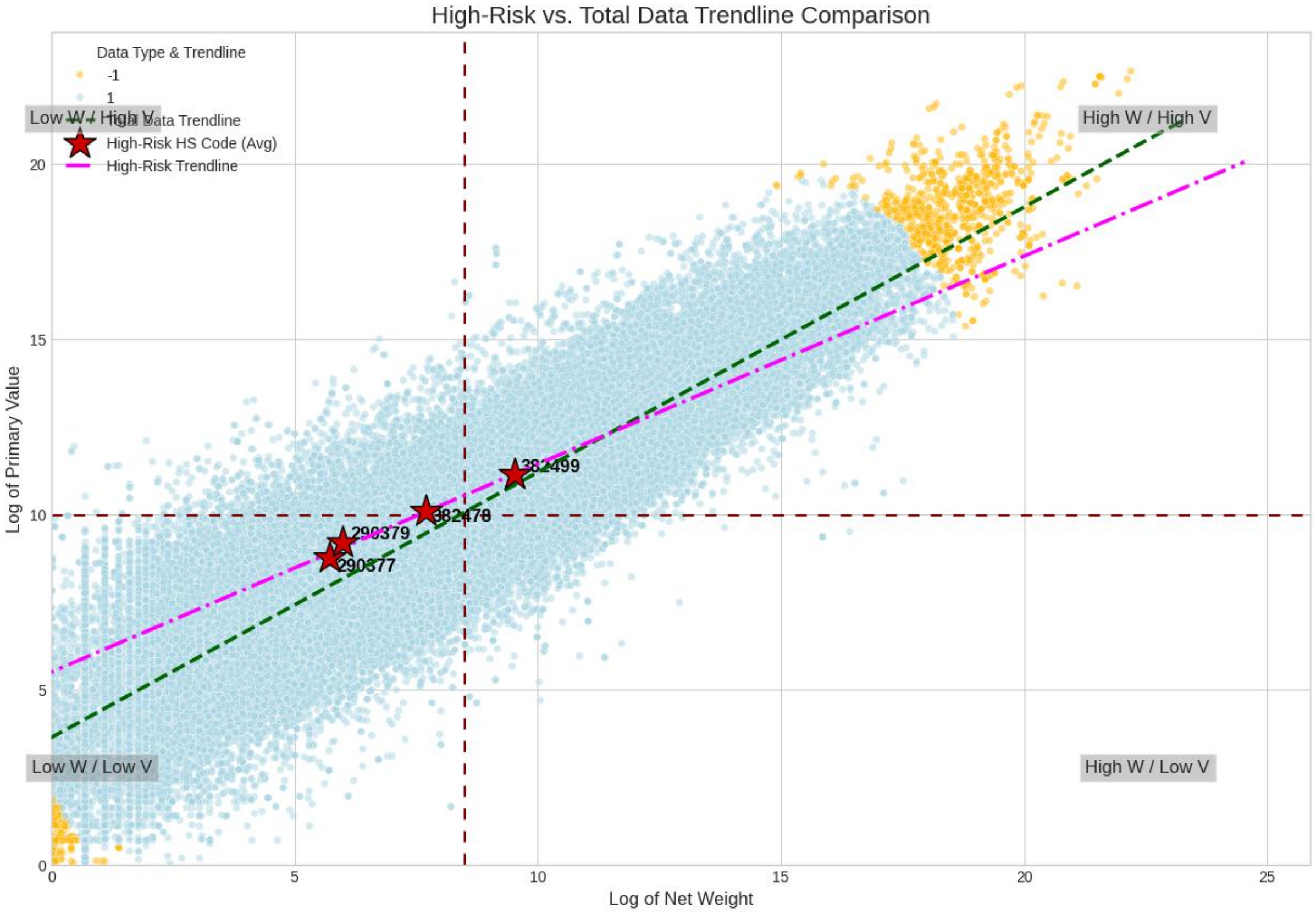

Source: Processed by Author (2025) using UN Comtrade Data (2020–2022)

The histogram in figure 5, visualizes the frequency of specific pricing anomalies discovered in trade data between these two locations. The horizontal X-axis represents the "Price per kg (USD)" spanning from 0 to over 4,000, while the vertical Y-axis indicates the "Frequency of Anomalous Trades," ranging from 0 to 25. The distribution is heavily skewed to the left. The most dominant feature is a single, tall purple bar located at the very beginning of the price scale (roughly 0 to 300 USD/kg). This indicates that the vast majority of flagged anomalous trades-nearly 25 instances-occur at this relatively low price point. A dark blue curve follows this trend, peaking sharply at the start and then flattening out as it moves to the right. In stark contrast to the main cluster, the chart reveals distinct, extreme outliers. There are two isolated, low-frequency bars located far to the right side of the graph: one appearing near the 2,700 USD mark and another exceeding 4,000 USD. These solitary bars represent rare but significant anomalies where the price per kilogram is exponentially higher than the typical anomalous trade on this route.

**Figure 5. Distribution of Anomalous Prices on a Key High-Risk Route.**

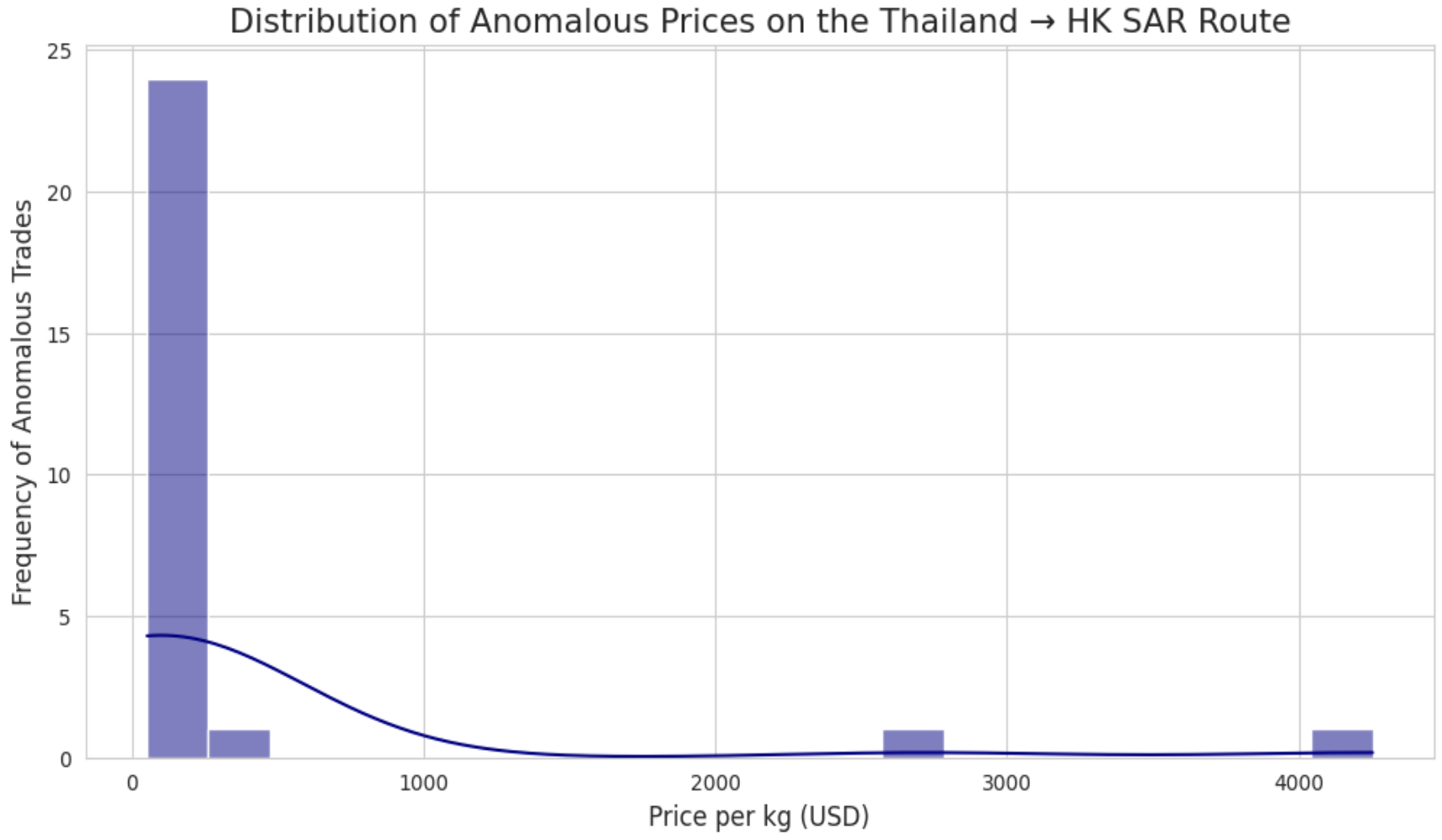

Source: Processed by Author (2025) using UN Comtrade Data (2020–2022)

**Figure 6. Maximum Import and SST Rates for ODS-related HS Codes in Malaysia.**

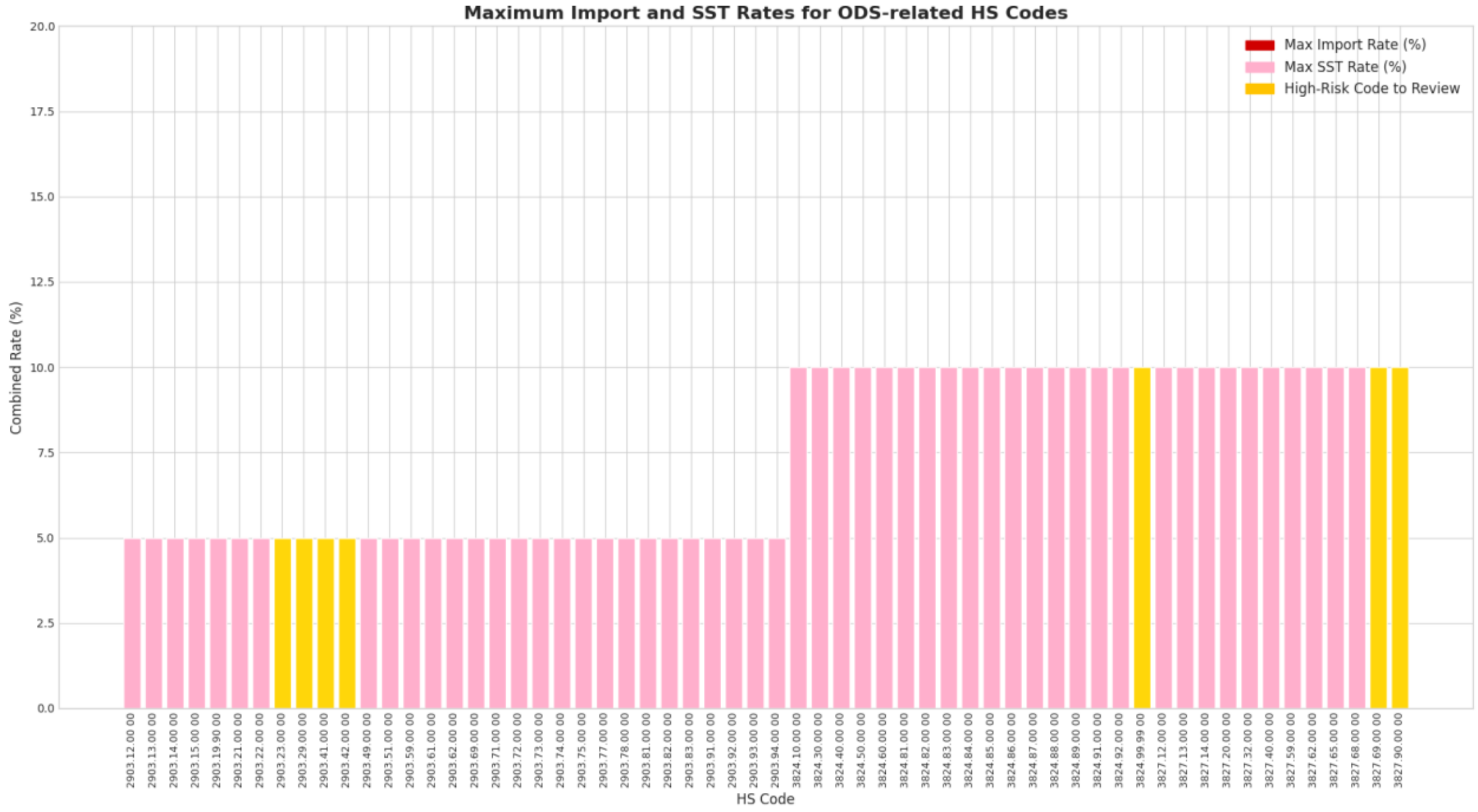

Source: Processed by Author (2025) using Malaysia Customs Duties Order 2025

The bar chart in figure 6 titled visualizes the applicable tax rates for a long sequence of specific Harmonized System (HS) codes, which are listed along the bottom horizontal axis. The vertical axis measures the "Combined Rate (%)" ranging from 0 to 20. The chart uses a color-coded legend to distinguish between different data points: pink bars represent the "Max SST Rate (%)," yellow bars indicate "High-Risk Code to Review". The data displays a distinct, two-tiered structure. The left half of the chart, comprising codes primarily beginning with "2903," shows a uniform flat rate of 5%. Roughly halfway through the chart, where the codes switch to the "3824" and "3827" series, the rate doubles, maintaining a consistent height of 10% for the remainder of the items. This visual step-change clearly separates the commodities into two different tax brackets based on their classification. Standing out against the rows of uniform pink bars are several yellow bars, which highlight specific codes flagged as high risk. In the lower 5% tax bracket, there is a cluster of three high-risk codes (2903.23, 2903.29, and 2903.41). In the higher 10% tax bracket, there is one isolated high-risk code (3824.99.99.00) followed by two more at the far-right end of the chart (3827.69 and 3827.90). These yellow highlights draw immediate attention to specific goods that require closer scrutiny, despite them sharing the same standard tax rates as the surrounding non-risky items.

### 3.2. Validation of Risk Drivers and Network Structures

To validate the factors driving the composite risk score, an Explainable AI (XAI) analysis using SHAP was performed. The feature importance plot, shown in Figure 7, provides a clear, interpretable quantification of each feature's contribution. This analysis confirmed that a high primaryValue and the is_vague flag (indicating a deliberately ambiguous commodity description) were the two most powerful predictors of a high-risk transaction. This finding is critical as it validates the framework's core hypothesis: that illicit trade patterns are frequently concealed within high-value shipments that use intentionally vague descriptions to avoid regulatory scrutiny. This validates the integration of the heuristic, knowledge-driven layer (Layer 3) with the purely statistical layers.

**Figure 7. SHAP Analysis of Feature Impact on Risk Score.**

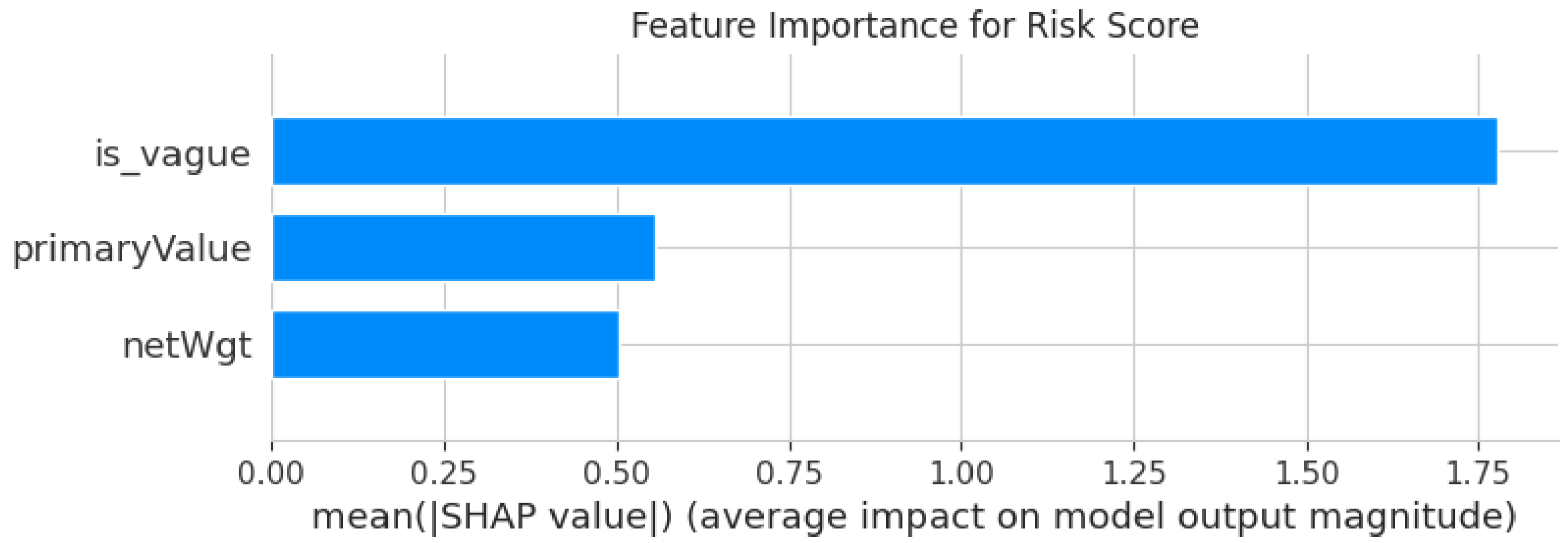

Source: Processed by Author (2025) using UN Comtrade Data (2020–2022)

This horizontal bar chart in figure 7 utilizes mean SHAP values to quantify the impact of three specific features on a predictive model's output. The horizontal X-axis measures the "mean (|SHAP value|)" or the average impact on the model output magnitude, with a scale ranging from 0.00 to roughly 1.80, while the vertical Y-axis lists the variables being analyzed. The chart highlights a clear and dramatic hierarchy of importance among the features. The variable is_vague is the overwhelmingly dominant factor, represented by a long bright blue bar that extends nearly to the 1.80 mark. This indicates that the presence of vague descriptions (likely related to the misclassification tactic mentioned in previous contexts) is the single strongest predictor within this risk model, vastly outperforming the quantitative metrics. In comparison, the remaining two features, primaryValue and netWgt (Net Weight), have a much smaller influence. Both appear with significantly shorter bars that hover just above the 0.50 mark, with "primaryValue" being slightly more impactful than "netWgt." This disparity demonstrates that while financial and physical dimensions are factors, the model heavily prioritizes the qualitative nature of the cargo description when assessing risk. Furthermore, analysis of the relational structure of high-risk trade revealed significant patterns of coordination. The unsupervised community detection analysis (using the Louvain method) identified four primary network patterns, or "trade blocs," within the high-risk data. These communities, which connected distinct regional clusters, such as jurisdictions in one continent with another, or countries within the same geographic region, suggest the existence of structured, regional smuggling networks rather than random, isolated incidents. This network analysis provides a crucial starting point for investigating potentially coordinated, multi-national illicit trade activities, transforming individual flagged shipments into a map of a larger illicit enterprise.

The scatter plot in figure 8 analyzes various global regions based on their position and behavior within a trade network. The horizontal X-axis measures "Network Centrality (Bridge Role)," ranging from 0.00 to 0.07, identifying how critical a location is as a connector in the supply chain. The vertical Y-axis tracks the "Import/Export Flow Ratio," where a value of 1.0 signifies a perfectly balanced trade flow. The chart represents each location as a bubble, where the size indicates the volume of total imports and the color represents the calculated risk index. The color spectrum runs from dark purple (low risk) to teal and green (moderate risk) up to bright yellow (high risk). The most dominant feature is a massive bright yellow bubble in the top-right corner, representing a major North American market. This indicates that this region has the highest network centrality, a balanced flow ratio, and the highest calculated risk index among the displayed data. The rest of the chart shows a clear divergence between major and minor trade hubs. Significant economies in East Asia and Western Europe appear as substantial green or teal bubbles in the center-right, indicating large import volumes and moderate-to-high centrality. Conversely, the left side of the plot is crowded with smaller, dark purple bubbles representing emerging markets in Southeast Asia, Eastern Europe, and Latin America. These regions display low network centrality (indicating they are not major bridges in this specific network) and a low-risk index, although their flow ratios vary significantly. A major Central European economy stands out as a unique outlier at the bottom center, showing moderate centrality but a very low import/export flow ratio.

**Figure 8. Transshipment Risk Profile (Centrality vs. Flow Ratio).**

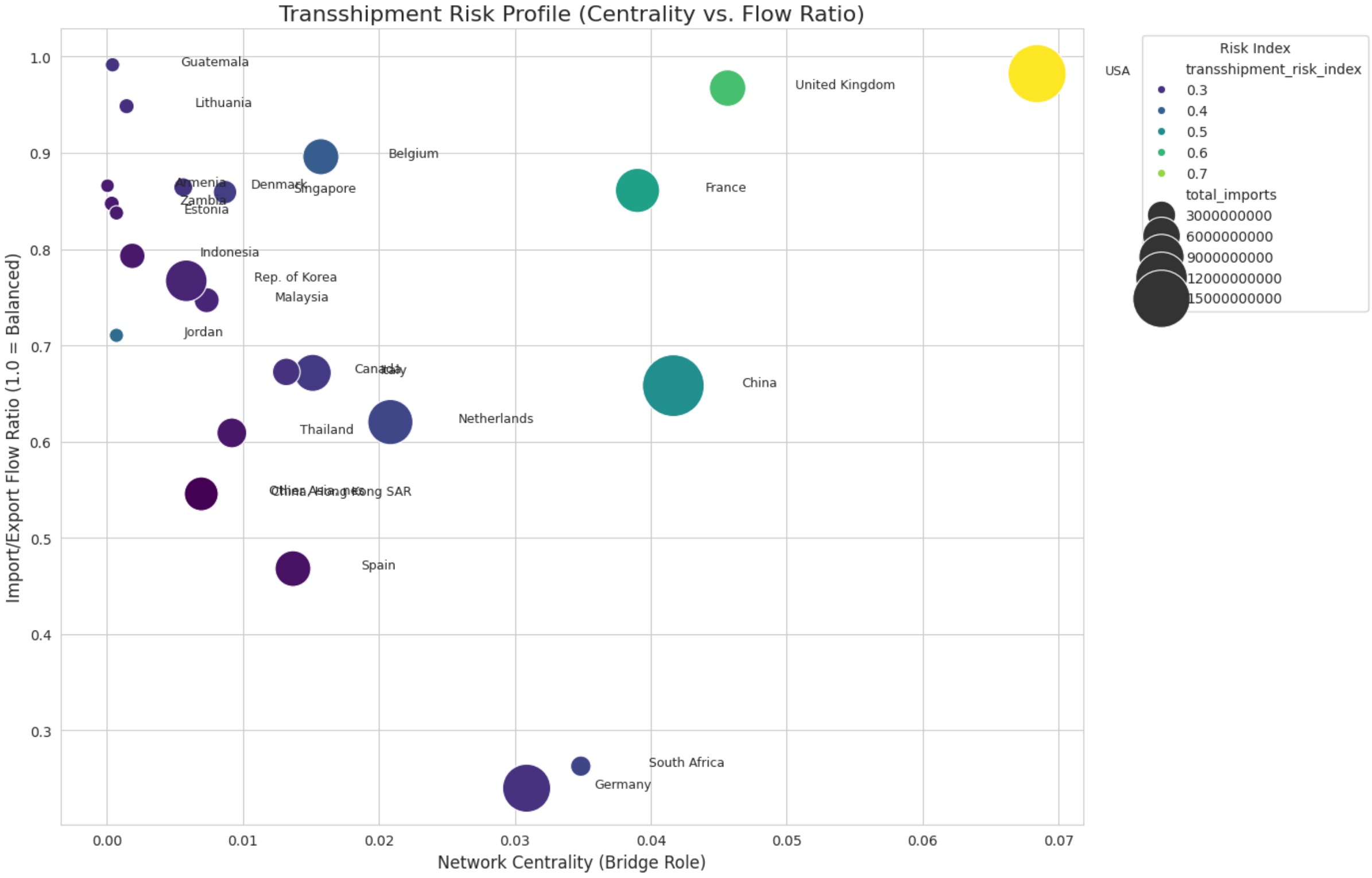

Source: Processed by Author (2025) using UN Comtrade Data (2020–2022)

**Figure 9. Blend Composition Word Cloud**

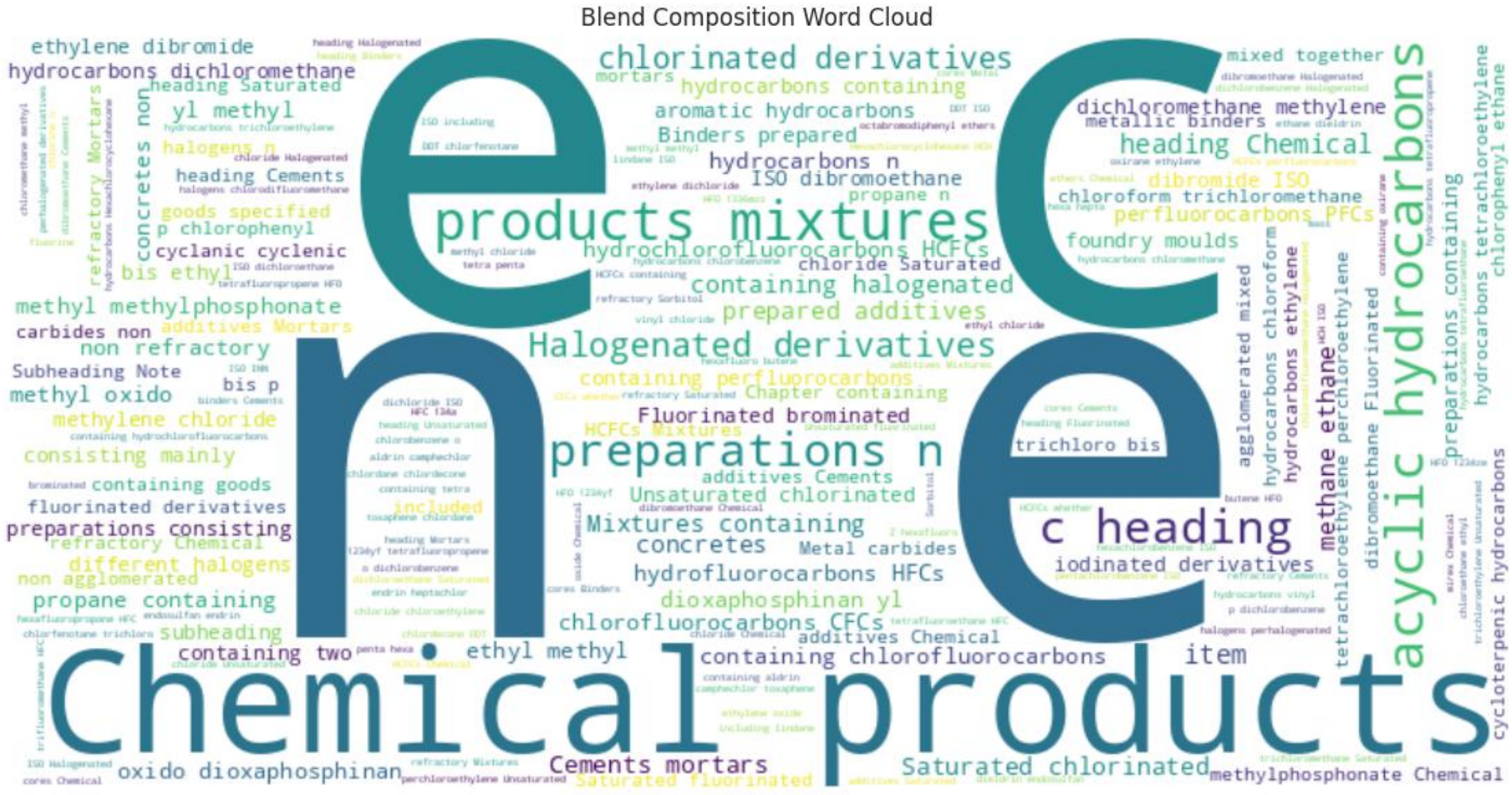

Source: Processed by Author (2025) using UN Comtrade Data (2020–2022)

The "Blend Composition Word Cloud" on figure 9 offers a powerful visual analysis of the textual data associated with flagged shipments. It is generated from the "generalized descriptions" of the transactions that the unsupervised learning framework isolated for customs review. The size of each word or phrase corresponds to its frequency within this high-risk dataset. The visualization confirms a key hypothesis of the abstract: anomalous shipments are often characterized by vague, "catch-all" descriptions, as evidenced by the prominence of terms like "n e c" (not elsewhere classified), "mixtures,"

"preparations," and "chemical products." The co-occurrence of these general terms with specific chemical keywords ("halogenated derivatives," "HCFCs," "chlorinated") reinforces the methodology of using "generalized descriptions" as a critical "heuristic flag" for developing the composite priority score.

### 3.3. Temporal Anomaly Detection: A Real-World Validation

Beyond identifying static routes, the framework's capacity for statistical anomaly detection using the Isolation Forest algorithm enabled a dynamic, temporal analysis of market-wide behavior. The isolation of "mega-trades" statistically rare outliers in value and volume allowed for the tracking of significant market shocks over time. As illustrated by the line charts on figure 10, an aggregated analysis of these mega-trades revealed a dramatic and anomalous peak in both the frequency and total value of such transactions occurring in early 2021. A geographical deconstruction of this spike, shown in the accompanying stacked area chart, identified the United States as the overwhelming driver of this surge. This algorithmically-detected spike correlates directly with a major external regulatory event: the enactment of the American Innovation and Manufacturing (AIM) Act on December 27, 2020. This landmark legislation empowered the U.S. Environmental Protection Agency to implement a steep, nationwide phasedown of hydrofluorocarbons (HFCs), creating a new allowance allocation and trading program that began in 2021. The resulting flurry of high-value transactions, as companies sought to secure allowances and position themselves within this newly regulated market, was successfully captured by our model as a massive statistical anomaly. This serves as a powerful validation of the framework's sensitivity, demonstrating its ability to transform raw trade data into actionable intelligence by detecting the real-world commercial impacts of major policy shifts, thereby providing a crucial tool for regulatory oversight.

**Figure 10. Trends in Global Mega-Trades (2020-2022), Detailing Overall Frequency and Value Against the Growing Contribution of Top Reporting Countries.**

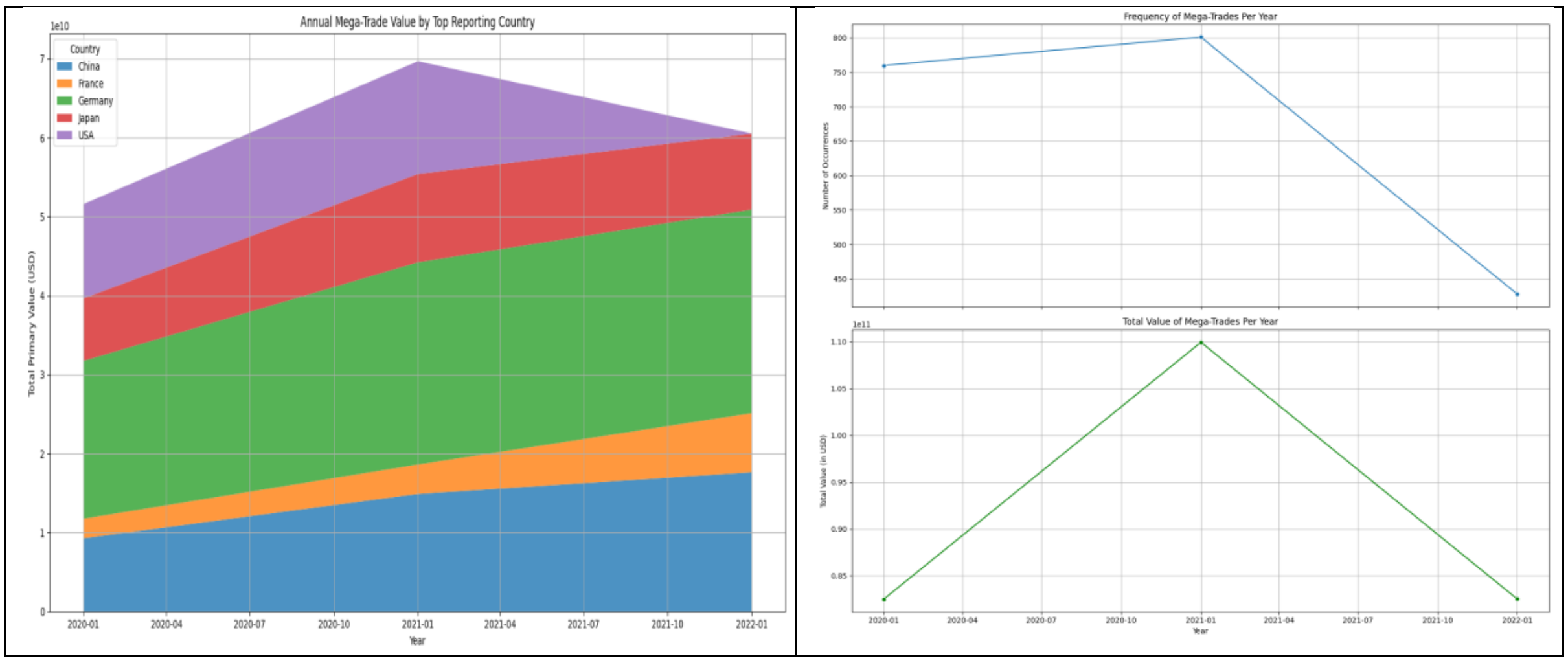

Source: Processed by Author (2025) using UN Comtrade Data (2020–2022)

The figure 10 presents a dashboard of three charts analyzing "Mega-Trade" patterns from early 2020 through the start of 2022. The largest visualization on the left is a stacked area chart titled "Annual Mega-Trade Value by Top Reporting Country," while the column on the right contains two-line charts tracking frequency and total value respectively. The main stacked area chart on the left illustrates the cumulative growth of trade value over time, broken down by major global regions. The layers represent contributions from key economies in East Asia, Western and Central Europe, and North America. The bottom blue and red layers correspond to East Asian markets, showing steady, consistent growth throughout the period. The middle orange and green sections represent European contributions, which expand significantly, particularly around 2021. The top purple layer represents the North American market, which shows a distinct bulge in volume during the middle of the timeline before tapering off slightly at the end. Overall, the total height of the chart peaks around the start of 2021, indicating a maximum cumulative trade value during that time. The two smaller line charts on the right provide a summary of the aggregate trends. The top chart, "Frequency of Mega-Trades Per Year," shows that the number of these large-scale transactions remained high and relatively stable between 2020 and 2021, peaking at around 800 occurrences, before experiencing a sharp precipitous drop to roughly 430 in 2022. The bottom chart, "Total Value of Mega-Trades Per Year," mirrors this trajectory but with a more pronounced spike. It shows that the total monetary value of these trades surged dramatically in 2021 to approximately 110 billion USD, up from around 83 billion USD in 2020, before falling back down to 2020 levels in 2022. Together, these charts suggest that 2021 was an outlier year characterized by both a high volume and high value of mega-trades.

## 4. Discussion

The ultimate objective of this forensic framework is to translate the recognized statistical patterns into prioritized, actionable intelligence for customs officials. This process directly aligns with the principles of intelligence-led enforcement, wherein data analysis serves as the crucial driver for allocating investigative resources effectively and proactively (Ratcliffe, 2016). The final intelligence package consists of several targeted outputs, each directly derived from the analytical layers of the framework, including comprehensive hotspot reports, geospatial risk maps, and prioritized case files for direct investigative use.

**Figure 11. Top 15 High-Risk Routes by Frequency.**

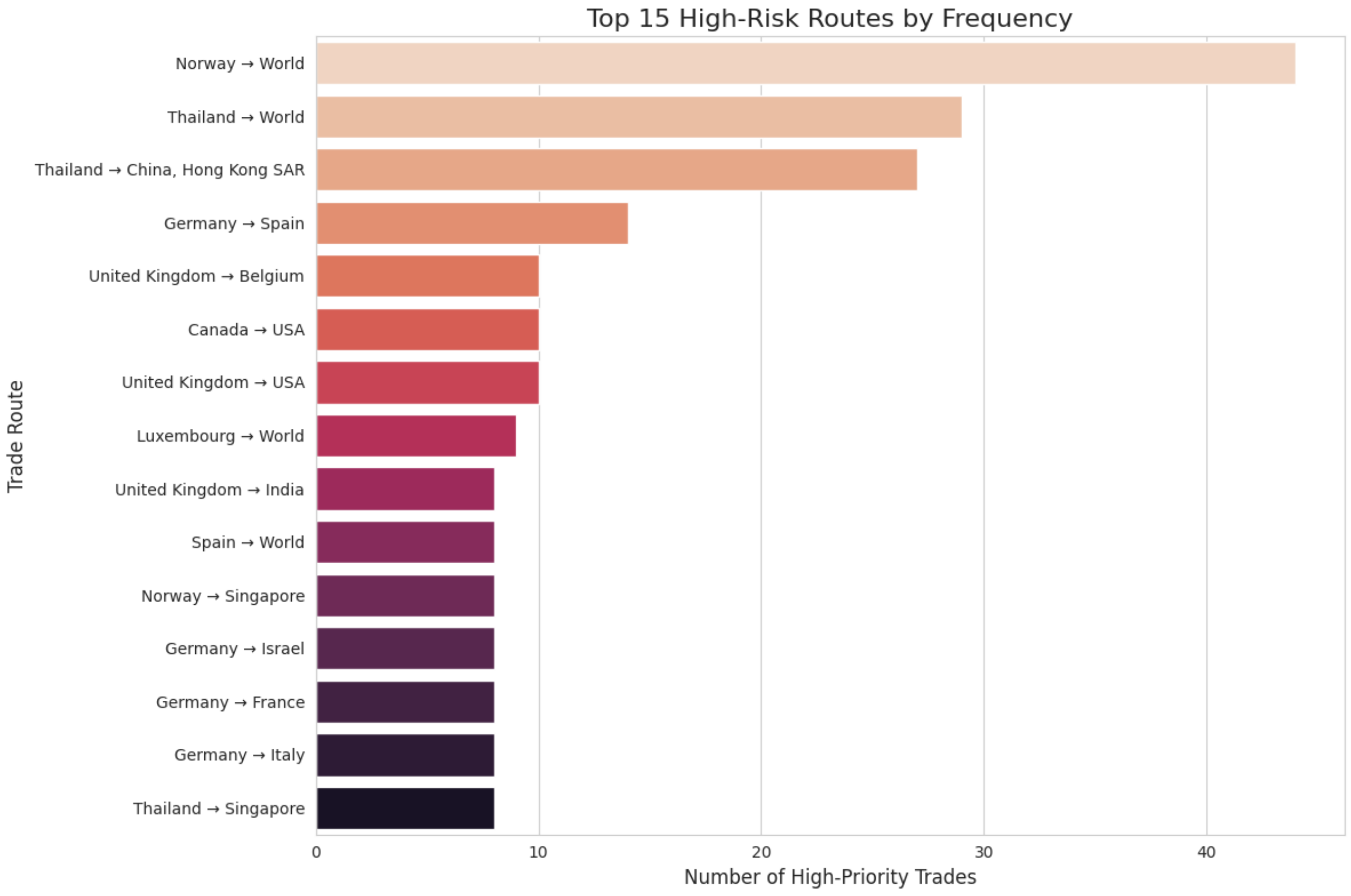

Source: Processed by Author (2025) using UN Comtrade Data (2020–2022)

The horizontal bar chart in figure 11, ranks specific trade corridors based on the volume of high-priority flagged trades. The horizontal axis measures the number of these trades, ranging from 0 to over 40, while the vertical axis lists the origin-to-destination pairings. The chart utilizes a color gradient that transitions from light beige at the top to dark purple at the bottom to visually distinguish the ranking order. The data is heavily skewed by the top three routes, which exhibit significantly higher frequencies than the rest of the group. The single most active high-risk route originates in Northern Europe and is destined for the global market ("World"), peaking at over 40 occurrences. This is followed by two major routes originating in Southeast Asia: one serving the global market and another targeting a specific major economy in East Asia. These three top entries are clearly separated from the rest of the pack, indicating primary hot spots for illicit or risky activity. Below this top tier, the frequency drops notably to a range of roughly 10 to 15 trades. This middle section is characterized by intra-regional connections within Central and Western Europe, as well as a prominent cross-border route within North America. The bottom half of the chart flattens out into a plateau, where numerous routes-involving exports from Central Europe to Southern Europe and the Middle East, as well as intra-regional trade in Southeast Asia-all show a consistent frequency of approximately 8 high-priority trades each.

**Figure 12. Top 15 Trading Entities by Transshipment Risk Index.**

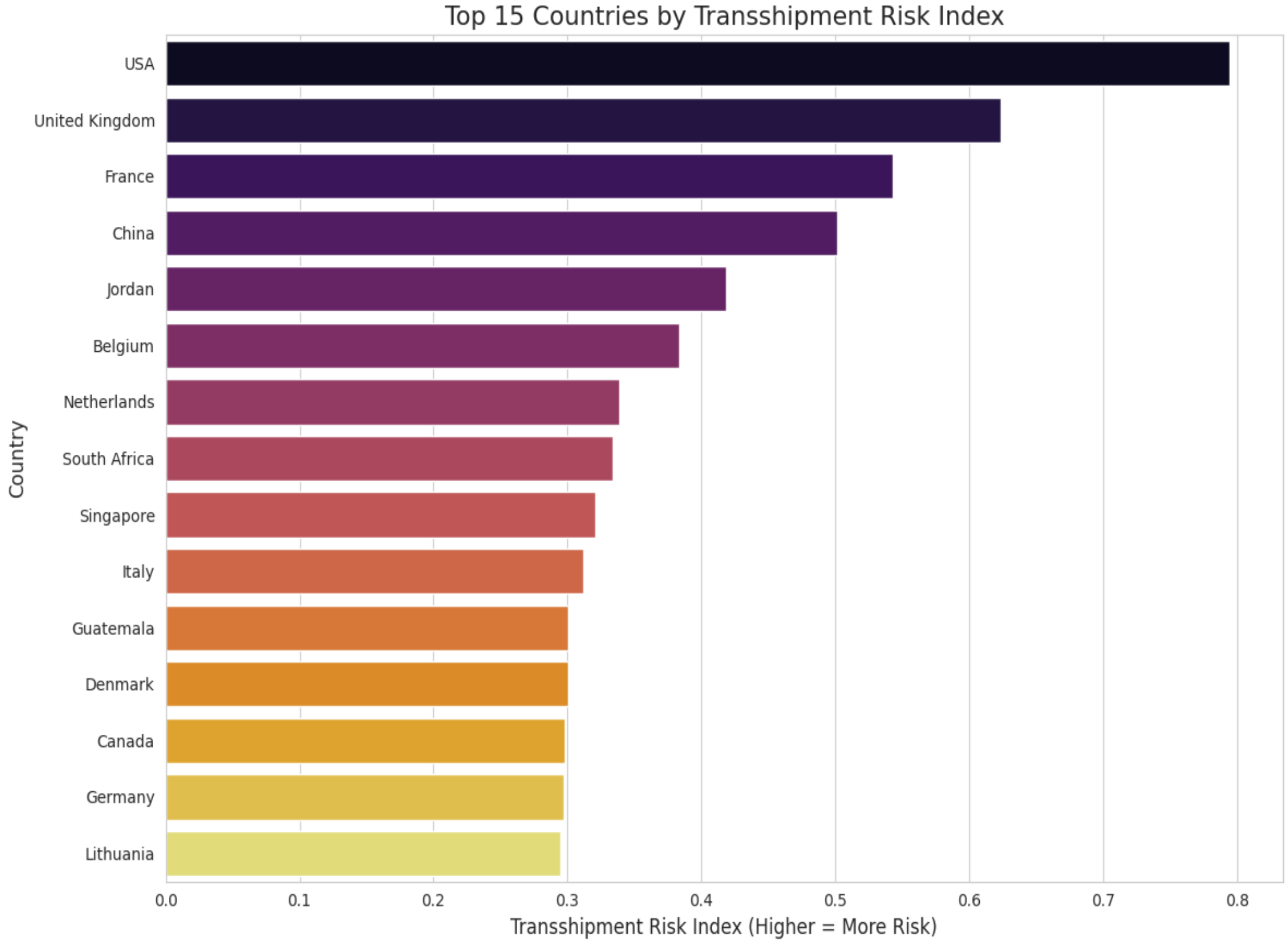

Source: Processed by Author (2025) using UN Comtrade Data (2020–2022)

This horizontal bar chart in figure 12 ranks specific nations based on their calculated risk levels within the transshipment network. The vertical axis lists the countries, while the horizontal axis measures the "Transshipment Risk Index," where a higher value indicates greater risk. The scale ranges from 0.0 to 0.8. The chart employs a color gradient that transitions from dark, near-black bars at the top to lighter yellow bars at the bottom, visually reinforcing the descent from highest to lowest risk among the top 15. The rankings are dominated by a major North American economy, which sits alone at the very top with a risk index nearing 0.8, significantly outpacing all other entries. It is followed by two major Western European powers and a dominant East Asian economy, which form a distinct second tier with indices ranging roughly between 0.5 and 0.65. This indicates that the largest global trade hubs are also the primary points of calculated risk in this specific model. Following this top tier, risk scores drop noticeably and begin to plateau. The middle of the chart features a diverse mix of regions, including the Middle East, Western Europe, and Southern Africa, with scores hovering between 0.35 and 0.45. The bottom third is tightly grouped around the 0.30 mark. This lower tier includes a wide geographic spread-spanning Southeast Asia, Southern Europe, Central America, Northern Europe, and the Baltic region-suggesting a consistent baseline level of risk across these secondary trade nodes.

The analysis first produced a ranked "Hotspot Report" of the most frequent high-risk trade routes, which is a direct application of pattern frequency analysis. As illustrated in Figure 11, "Top 15 High-Risk Routes by Frequency," the routes identified based on the high number of recurring anomalies were one specific jurisdiction to World and another major export hub to World. A more detailed network view, visualized in the figure 17, provides clearer, actionable intelligence by pinpointing the most critical specific trade channels. This flow analysis confirms a specific high-volume trade corridor as the dominant hotspot, representing the most frequent channel for high-risk ODS-related trades found in the dataset. Furthermore, this diagram reveals that this particular export hub acts as a major source for suspicious goods to a diverse set of partners, including major economies in North America, Asia, and Europe. While these reports focus on frequency and network flow, the heatmap analysis in Figure 15 identifies hotspots by total transaction value. This analysis pinpoints other highly concentrated routes, such as those originating from another major economy to partners including jurisdictions in the Middle East and the Americas, as well as another significant channel within a major European economic bloc between two neighbouring member states.

**Figure 13. Global Heatmap of Average ODS Trade Risk Score.**

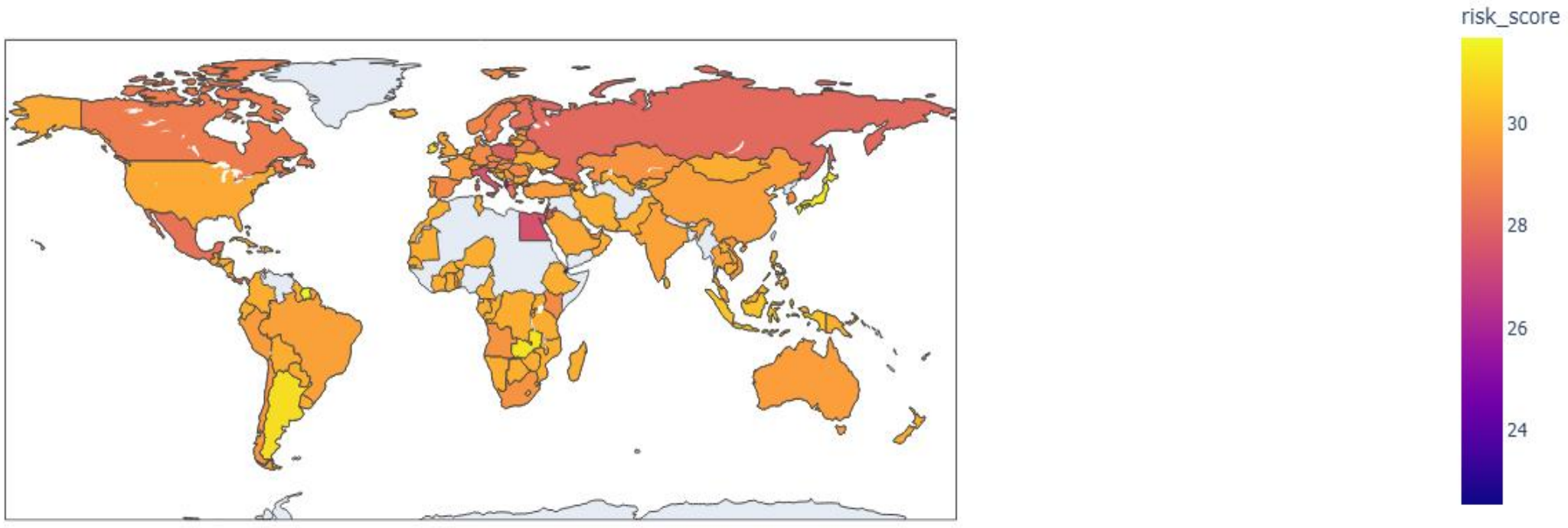

Source: Processed by Author (2025) using UN Comtrade Data (2020–2022)

An interactive choropleth map displaying the average risk score for each reporting country, with a color scale from yellow (low risk) to red (high risk) indicating global hotspots. (Note: A static screenshot of the interactive output would be used in the paper). Complementing this route-specific analysis, the framework generated a geospatial intelligence overview in the form of an interactive choropleth world map, seen in Figure 13. This tool provides a clear, high-level visualization of risk concentration by identifying countries with the highest average risk score across all their export transactions, allowing enforcement agencies to quickly identify national-level hotspots. The intelligence package also serves to validate the heuristic flagging methodology. As shown in the bar chart in figure 18, the most frequently flagged shipments, particularly from certain major economies, consistently use vague and complex commodity descriptions. This includes terms such as 'Mixtures containing halogenated derivatives' and 'not elsewhere specified or included’, confirming that this pattern of intentional obfuscation is a key feature used to prioritize entries for the final case file. The most critical output for operational enforcement is "The Final Case File," an exportable CSV file containing a "Highest Priority" list. This file provides a clean, ready-to-investigate dataset for customs officials, specifically containing the 1,288 entries that exhibited both a significant price anomaly and a vague commodity description. Finally, these complex findings were synthesized into a strategic, non-technical "Policy Memo". An officer would receive the prioritized list of 1,288 entries. They could then query these specific shipment IDs in their internal system, flag them for physical inspection upon arrival, or use the list to open a deeper investigation into the specific importers and exporters involved. This automatically generated document identifies key exporter hotspots (including several major European and Asian economies) and key destination hubs (including major economies in Europe and North America), quantifies the total value of flagged trade (over $182 billion), and provides clear recommendations for targeted audits and bilateral engagement.

**Figure 14. Total Trade Value of ODS-Flagged Entries Over Time**

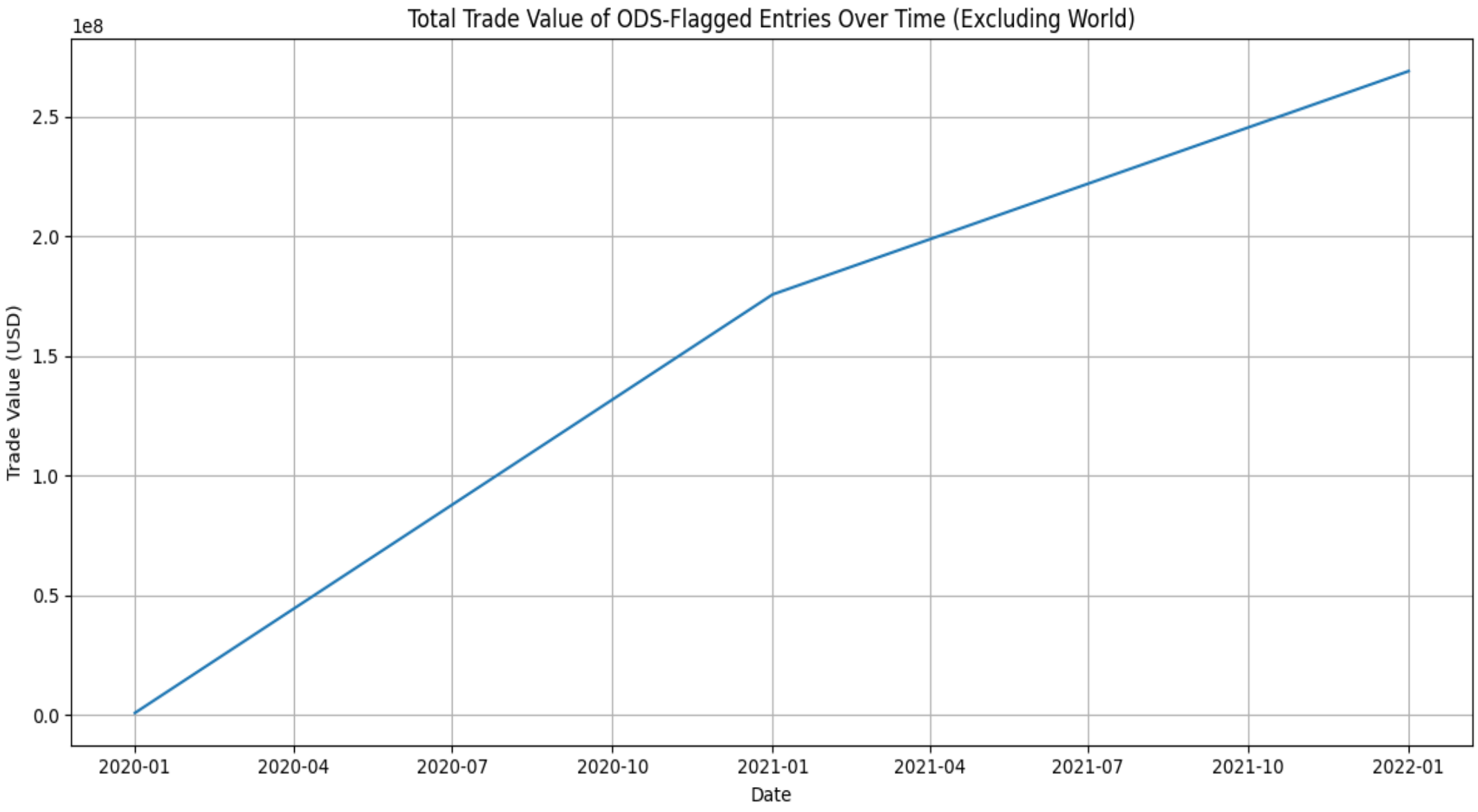

Source: Processed by Author (2025) using UN Comtrade Data (2020–2022)

This line chart in figure 14 illustrates the cumulative financial growth of specific flagged trade activities over a two-year period, spanning from January 2020 to January 2022. The vertical axis measures the "Trade Value (USD)" with a scale in hundreds of millions ($10^8$), while the horizontal axis marks the progression of time in quarterly intervals. The trendline depicts a continuous and significant upward trajectory across the entire timeline. Starting from a value near zero at the beginning of 2020, the trade value experiences a sharp and steady ascent. By the one-year mark in January 2021, the total flagged value has already surged to approximately 175 million USD, indicating a period of rapid accumulation. This growth persists through the second year, although the slope of the line softens slightly, suggesting a marginally slower but still consistent rate of increase compared to the first year. By the final data point in January 2022, the total value peaks at roughly 270 million USD. The chart confirms that the financial volume of these specific flagged entries has expanded substantially and without interruption over the observed period.

**Figure 15. ODS-Flagged Trade Value by Country Pair.**

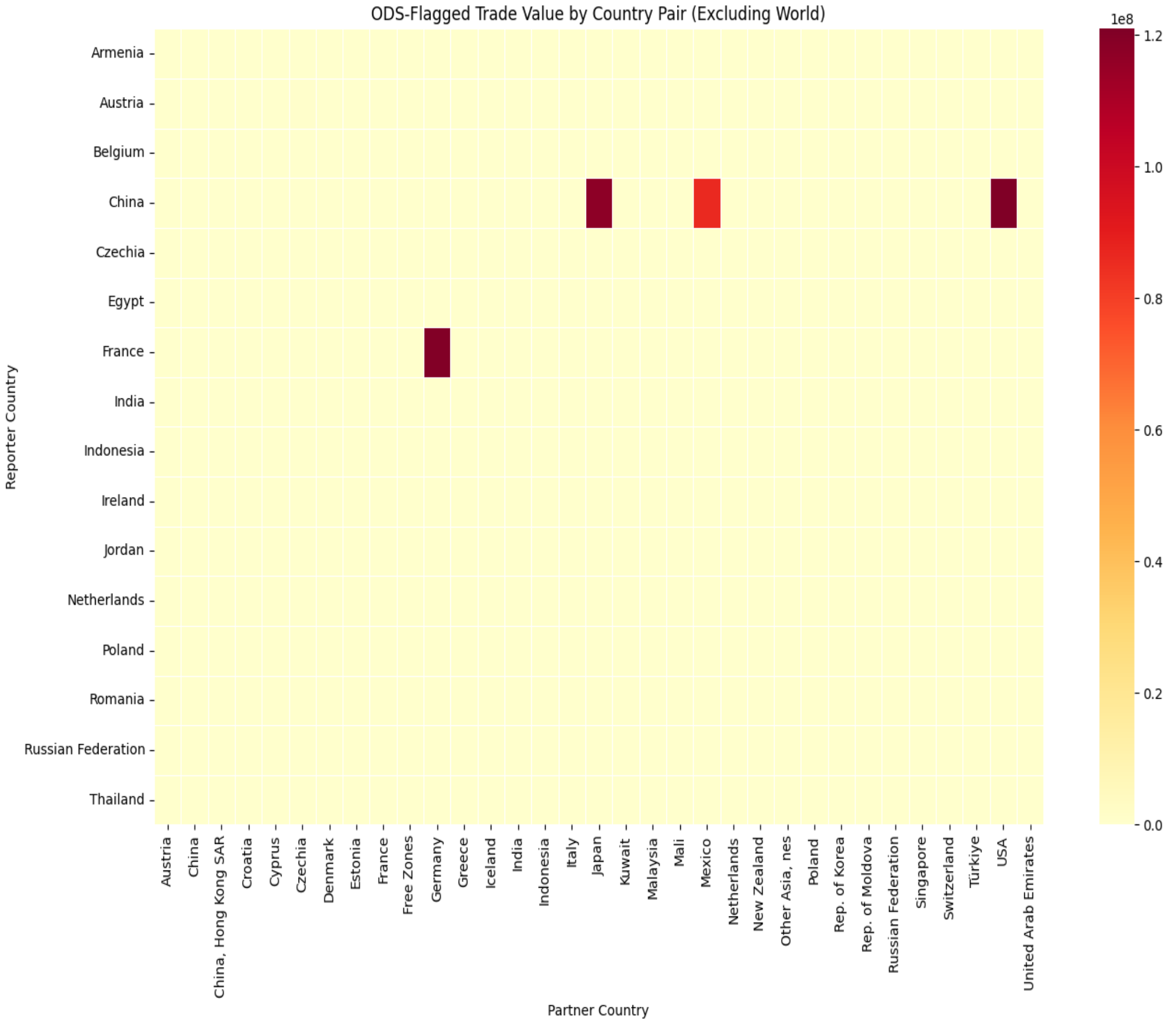

Source: Processed by Author (2025) using UN Comtrade Data (2020–2022)

The heatmap matrix in figure 15 visualize the financial intensity of trade between specific reporting locations (listed on the vertical Y-axis) and their partners (listed on the horizontal X-axis). The chart relies on a color scale found on the right, ranging from pale yellow (representing zero or negligible value) to deep red (representing a maximum value of approximately 120 million USD). The overwhelming prevalence of the pale-yellow color across the grid immediately suggests that for the vast majority of possible pairings, there is no significant flagged trade activity. However, the chart reveals a few highly specific, isolated corridors of intense activity. The most active "Reporter" is a major economy in East Asia, which is responsible for three of the four visible hotspots. This region shows a deep red, high-value connection with a neighboring East Asian market, as well as another deep red connection with a primary North American market. A third, slightly lighter orange-red square indicates a substantial trade flow between this same East Asian reporter and a secondary North American partner. Outside of this East Asian cluster, there is only one other significant outlier. A reporting nation in Western Europe displays a solitary, high-intensity dark red block, indicating a massive volume of flagged trade with a neighboring Central European partner. This visual isolation underscores that the high-risk trade value is not spread evenly across the global network but is instead highly concentrated within just a few specific bilateral relationships involving key economic hubs in Asia, North America, and Europe.

**Figure 16: Top ODS-Related Commodity Descriptions**

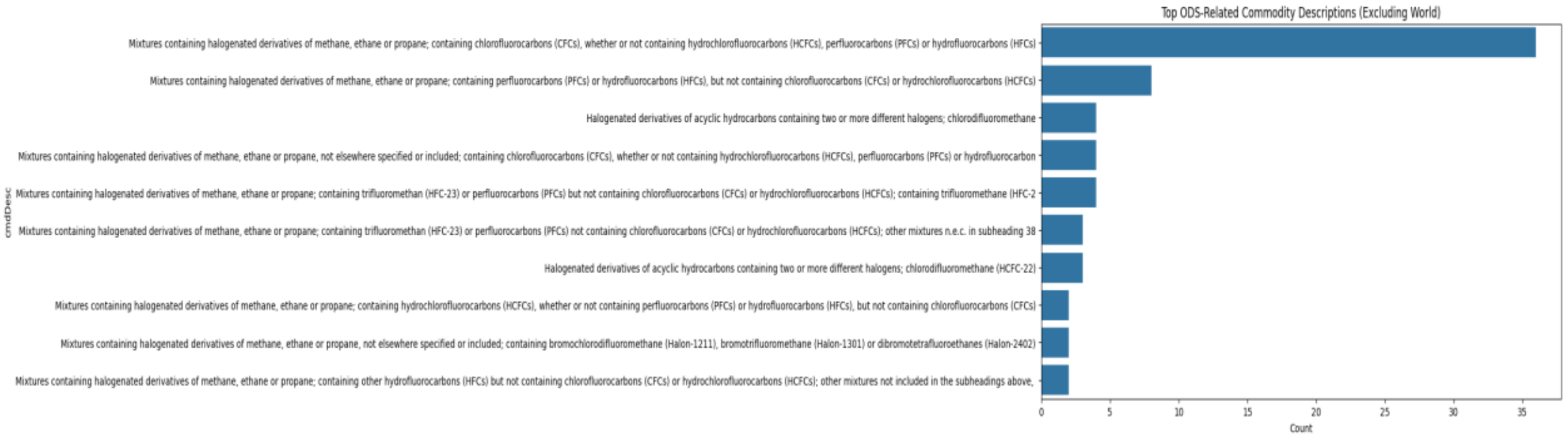

Source: Processed by Author (2025) using UN Comtrade Data (2020–2022)

This horizontal bar chart in figure 16 it illustrates the frequency of specific text descriptions used in trade documentation for Ozone Depleting Substances (ODS). The vertical axis lists long, technical commodity descriptions, which are primarily variations of "Mixtures containing halogenated derivatives of methane, ethane or propane" and specify the inclusion or exclusion of various chemicals like chlorofluorocarbons (CFCs), hydrochlorofluorocarbons (HCFCs), and hydrofluorocarbons (HFCs). The horizontal axis measures the "Count," indicating how often each specific description appears in the dataset. The data reveals a single, overwhelmingly dominant category at the very top. This description, which appears to be a broad classification for mixtures containing these halogenated derivatives, extends past the 30 marks on the count scale, dwarfing all other entries. The second most frequent description appears significantly less often, with a count of approximately 8. The remaining categories show a gradual decline in frequency, forming a "long tail" distribution where most specific descriptions appear fewer than 5 times. This suggests that while there are many specific ways to classify these goods, a single, likely broader description is used for the vast majority of these specific trade entries.

**Figure 17. Flow Diagram of High-Risk ODS Trade Routes.**

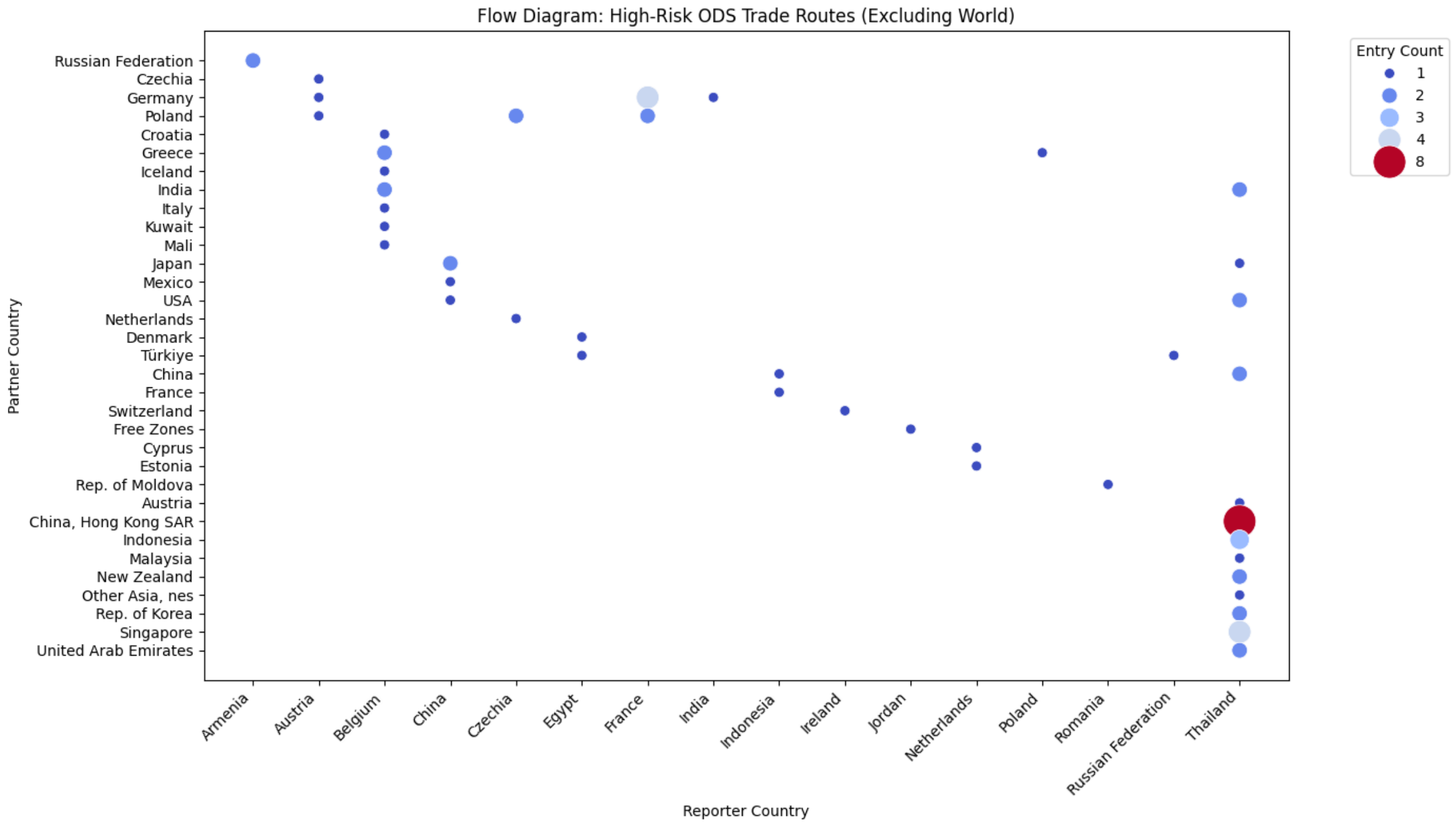

Source: Processed by Author (2025) using UN Comtrade Data (2020–2022)

The scatter plot in figure 17 visualizes the network of flagged trade connections between reporting locations on the horizontal axis and partner destinations on the vertical axis. The chart uses bubble size and color to represent the "Entry Count," with small blue dots indicating low-frequency connections (1 to 4 trades) and a large red dot representing the highest frequency (8 trades). The most striking feature of the diagram is the intense activity associated with a single reporting jurisdiction located in Southeast Asia, positioned at the far right of the horizontal axis. This column is densely populated with vertical data points, indicating that this specific location is reporting high-risk trade with a diverse array of global partners, ranging from neighbors in Southeast Asia and East Asia to nations in Europe and the Middle East. Dominating this active column-and indeed the entire chart-is a large red bubble. This marker highlights a specific, high-frequency trade route connecting the Southeast Asian reporter to a major logistics hub in East Asia. This single connection represents the highest concentration of risky entries in the dataset. In contrast, the rest of the chart shows much sparser activity, with reporters in Western Europe, Eastern Europe, and North Africa displaying only scattered, low-frequency blue dots connecting to various partners.

**Figure 18. Top 15 Reporter Countries by Vague Description Frequency.**

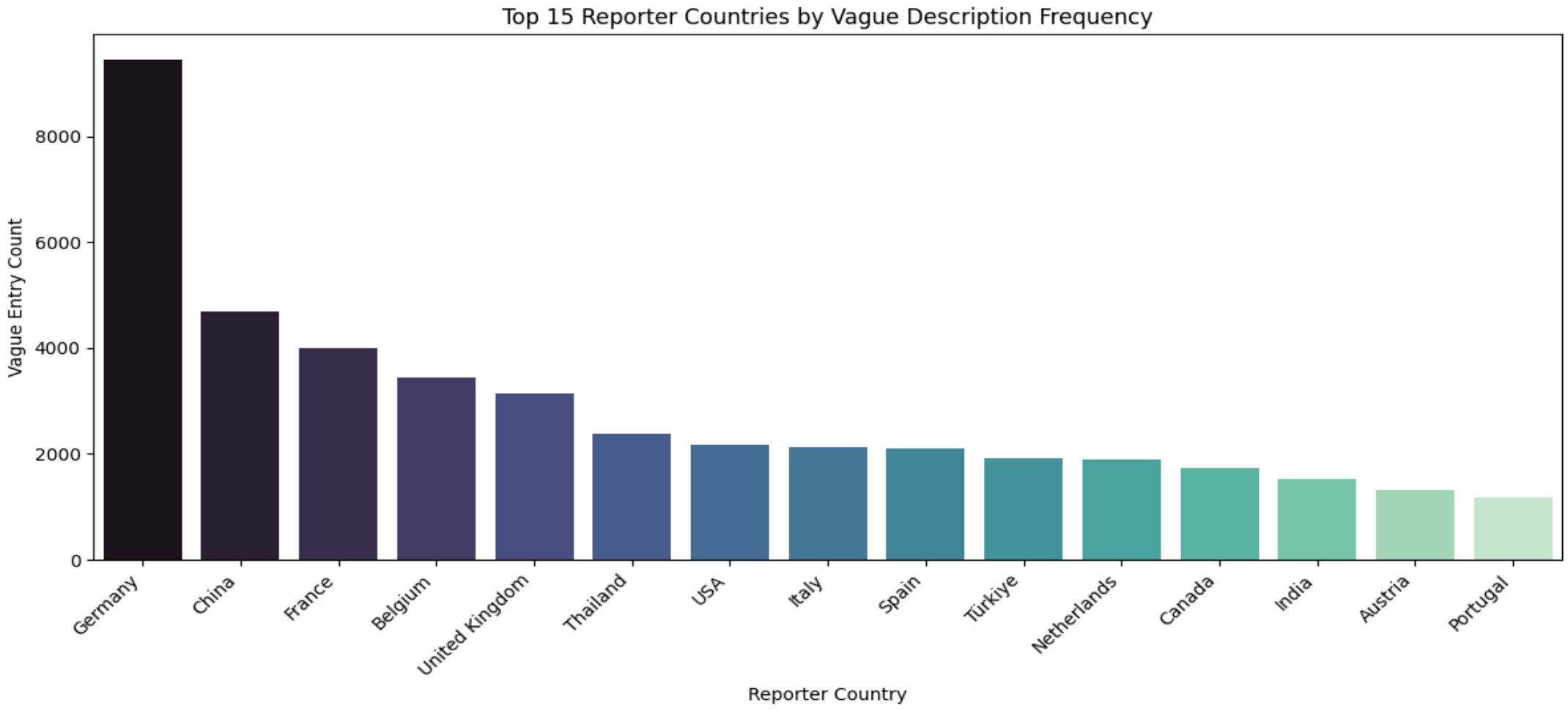

Source: Processed by Author (2025) using UN Comtrade Data (2020–2022)

The vertical bar chart in figure 18 ranks reporting jurisdictions based on how often they utilize non-specific labelling in their trade data. The vertical axis tracks the "Vague Entry Count," ranging from 0 to over 9,000, while the horizontal axis lists the reporting locations. The visuals employ a color gradient that shifts from dark purple-black on the left to pale mint green on the right, emphasizing the decreasing volume of vague entries across the ranking. The data is dominated by a massive outlier at the number one position. A major Central European economy stands alone with a vague entry count exceeding 9,000, which is roughly double the volume of the next highest entry. Following this peak, there is a significant drop-off to a second tier of reporters, comprising a dominant East Asian economy and several key Western European nations, which record frequencies ranging between 3,000 and 5,000. The remainder of the chart illustrates a gradual levelling off. A diverse middle group-including a Southeast Asian market, a major North American economy, and several Southern European nations-clusters tightly around the 2,000 marks. The distribution then tapers gently toward the right, with smaller economies in South Asia and Europe showing the lowest frequencies of vague descriptions among the top 15.

**Figure 19. Top 15 High-Risk ODS Trade Flows.**

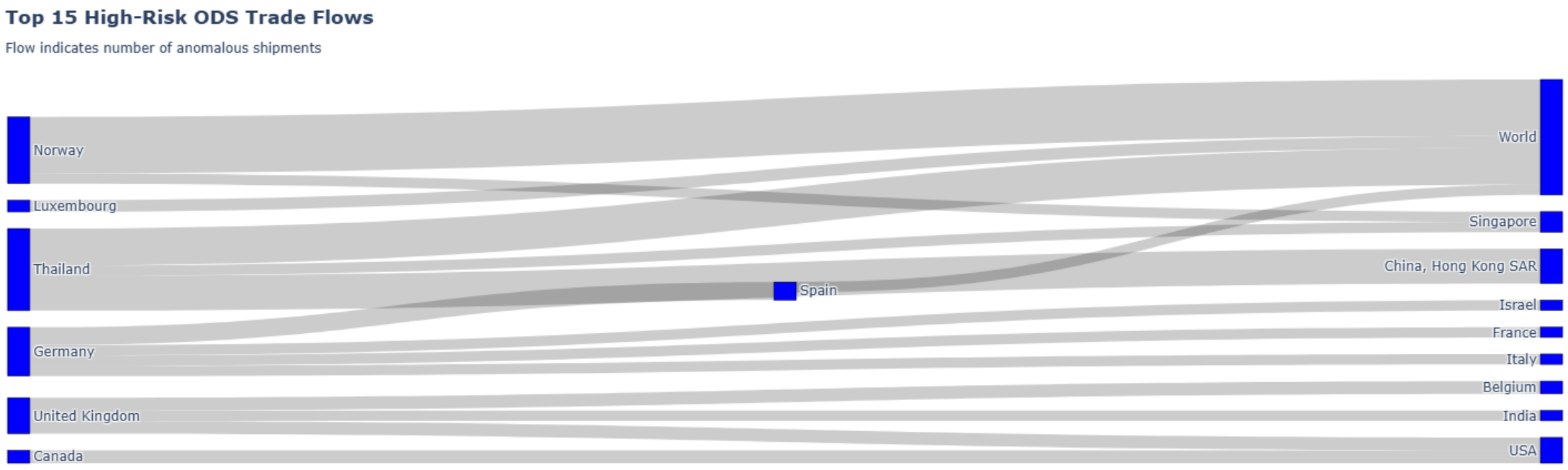

Source: Processed by Author (2025) using UN Comtrade Data (2020–2022)

The Sankey diagram, "Top 15 High-Risk ODS Trade Flows," provides a dynamic visualization of the anomalous trade network identified by the framework. The nodes on the left represent the originating entities of shipments, while the nodes on the right represent the destination entities. As indicated by the subtitle, the thickness of each flow corresponds directly to the "number of anomalous shipments" flagged by the unsupervised learning modules (such as the Isolation Forest or IQR price outlier detection). Instead of just listing high-risk entities, this chart maps the specific, high-priority trade corridors being used for these suspect transactions, transforming the "prioritized list of shipments" from the abstract into a clear relational map for regulatory enforcement.

**Figure 20. Top 15 HS Codes by Average Risk Score.**

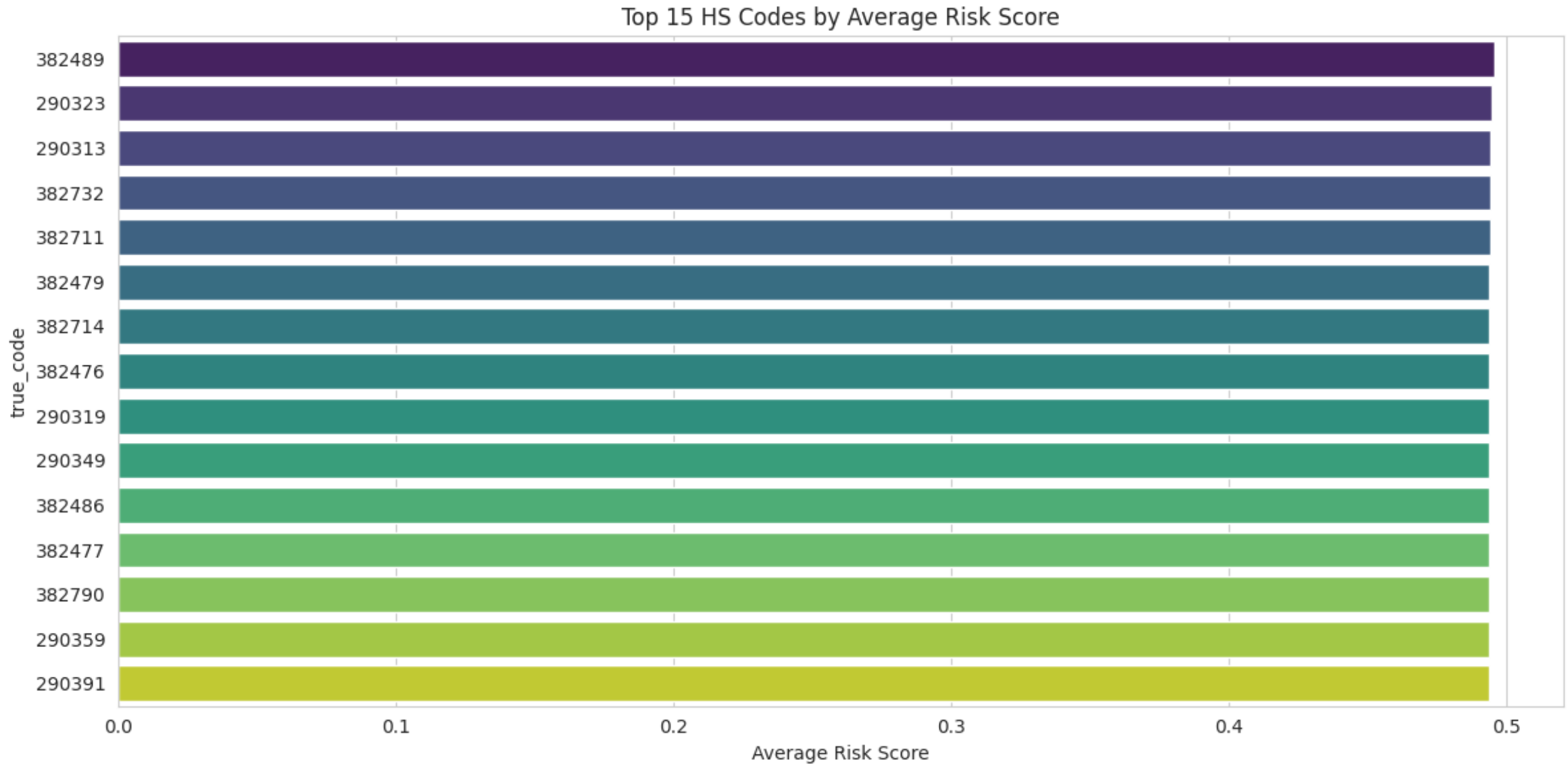

Source: Processed by Author (2025) using UN Comtrade Data (2020–2022)

The bar chart in figure 20 presents a prioritized list of high-risk *commodity codes* as identified by the framework. The vertical axis displays the top 15 Harmonized System (HS) codes, while the horizontal axis quantifies their "Average Risk Score." This score is an aggregated metric, likely representing the average "composite priority score" of all individual shipments classified under each specific code. This visualization pinpoints the exact product categories that are most frequently associated with the anomalous patterns (such as price-per-kilogram outliers or 'mega-trades') detected by the unsupervised learning modules. It provides highly specific, actionable intelligence, enabling regulatory agencies to focus their enforcement efforts on the particular goods that are most consistently flagged as high-risk.

## 5. Conclusion and Future Work

This study successfully demonstrates that a multi-modal unsupervised pattern recognition framework can effectively identify, profile, and prioritize suspicious trades concealed within large and noisy customs datasets. The combination of unsupervised clustering to establish a baseline of "normal" trade archetypes, layered anomaly detection (both statistical and commercial) to find significant deviations, and heuristic flagging to capture domain-specific knowledge proved highly effective in triaging the data. A novel finding was the discovery of a trendline divergence pattern between general goods and high-risk commodities; this pattern provides a new, quantifiable indicator for suspicion by showing that high-risk goods have a fundamentally different value-to-weight ratio. This work presents a replicable pipeline for transforming raw data into the actionable intelligence required for modern, data-driven regulatory enforcement. However, the analysis has limitations that must be acknowledged. The framework relies on publicly available data, which may contain inherent inaccuracies or reporting gaps. Furthermore, the interpretation of "risk" is derived from unsupervised learning, meaning it is primarily statistical and heuristic rather than a definitive confirmation of illicit activity. This necessitates a final validation step by domain experts in customs and chemical regulations, a crucial stage in the data science lifecycle to ensure that statistical findings translate into sound operational intelligence (Provost & Fawcett, 2013). The risk score is intended as a 'first-pass' filter for expert human review, not as a final automated decision, thereby mitigating the risk of disrupting legitimate commerce. Future work will focus on the full operationalization of several advanced machine learning models that were experimentally validated during the broader scope of this research. While the core framework presented in this paper relies on a validated and computationally efficient pipeline, our broader research included preliminary implementations of cutting-edge deep learning techniques that confirmed their significant potential for enhancing this framework. For textual analysis, experimental fine-tuning of a BERT (Bidirectional Encoder Representations from Transformers) model for HS code classification demonstrated a superior capability for understanding contextual nuance in commodity descriptions, validating its potential to drastically improve the detection of textual obfuscation (Devlin et al., 2019). For anomaly detection, a more sophisticated, deep-learning-based unsupervised method was successfully implemented using Variational Autoencoders (VAEs), which proved highly effective at identifying subtle anomalies based on reconstruction probability in complex datasets (An & Cho, 2015). Finally, to move beyond simple community detection and analyze the dynamics of trade networks, a Graph Neural Network (GNN) was used to model complex relational patterns like risk propagation, an approach that has shown success in analogous forensic domains such as anti-money laundering (Weber et al., 2019). The output of this experimental GNN model is visualized in Figure 21. This bar chart acts as a proof-of-concept, providing a prioritized list for regulatory review by ranking entities based on a composite score. This advanced score integrates an entity's network centrality its structural importance as a trade hub with the various risk indicators identified by the unsupervised learning framework. The horizontal axis quantifies this composite score, allowing for a clear, data-driven ranking and pinpointing the key entities whose combination of high-risk trade patterns and significant network position warrants further investigation. The successful preliminary implementation of these models provides a strong, evidence-based justification for their inclusion in future iterations of this forensic framework. The next stage of research should therefore focus on refining, scaling, and formally integrating these advanced components to build a next-generation intelligence suite for regulatory enforcement.

**Figure 21: Top 15 Trading Entities by GNN-Based Risk/Centrality Score.**

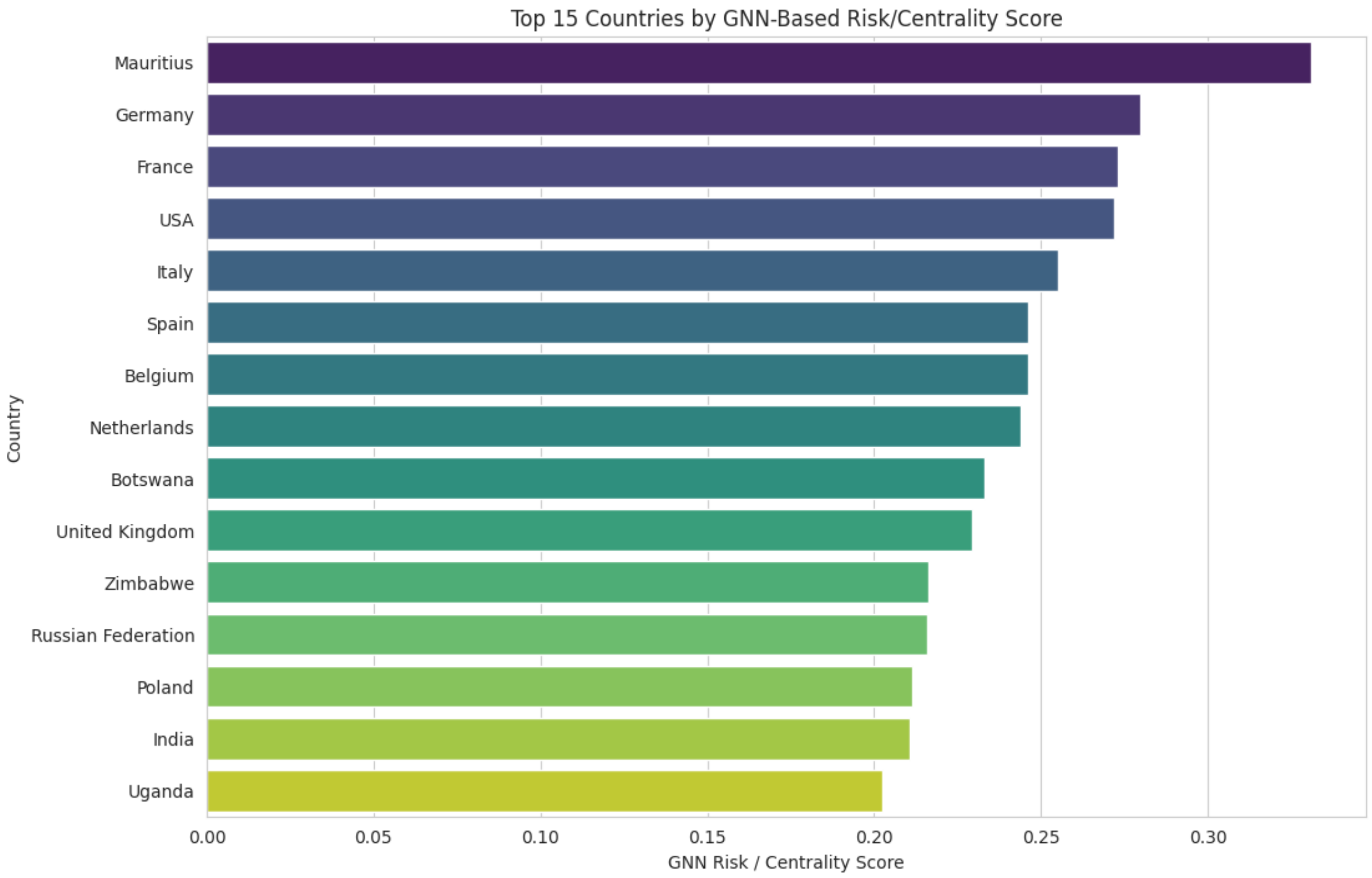

Source: Processed by Author (2025) using UN Comtrade Data (2020–2022)